\documentclass[runningheads]{llncs}
\usepackage{graphicx}
\usepackage{tikz}
\usepackage{comment}
\usepackage{amsmath,amssymb} % define this before the line numbering.
\usepackage{color}

\usepackage[accsupp]{axessibility}  % Improves PDF readability for those with disabilities.

\usepackage{makecell}
\usepackage{graphicx}
\usepackage{amsmath, bm}
\usepackage{tabularx,booktabs}
\newcolumntype{C}{>{\centering\arraybackslash}X} % centered version of "X" type
\usepackage{adjustbox}
\usepackage{multirow}
\usepackage{graphicx,xcolor,colortbl} 
\definecolor{LightCyan}{rgb}{0.88,1,1}
\usepackage{marvosym}
\usepackage{appendix} %%!!!!!!!!

\begin{document}
% \linenumbers
\pagestyle{headings}
\mainmatter
\def\ECCVSubNumber{1928}  % Insert your submission number here

\title{BinsFormer: Revisiting Adaptive Bins for Monocular Depth Estimation} % Replace with your title

\begin{comment}
\titlerunning{ECCV-22 submission ID \ECCVSubNumber} 
\authorrunning{ECCV-22 submission ID \ECCVSubNumber} 
\author{Anonymous ECCV submission}
\institute{Paper ID \ECCVSubNumber} 
\end{comment}
%******************

% \begin{comment}
\titlerunning{BinsFormer}
\author{Zhenyu Li\inst{1} \and
Xuyang Wang\inst{2} \and
Xianming Liu\inst{1} \and
Junjun Jiang\inst{1}\textsuperscript{\Letter}}
\authorrunning{Z. Li et al.}
\institute{Harbin Institute of Technology \and
Australian National University \\
\email{\{zhenyuli17,csxm,jiangjunjun\}@hit.edu.cn xuyang.wang@anu.edu.au}}

% \end{comment}
%******************
\maketitle 

\begin{abstract}

Monocular depth estimation is a fundamental task in computer vision and has drawn increasing attention. Recently, some methods reformulate it as a \textit{classification-regression} task to boost the model performance, where continuous depth is estimated via a linear combination of predicted probability distributions and discrete bins. In this paper, we present a novel framework called \textbf{BinsFormer}, tailored for the \textit{classification-regression-based} depth estimation. It mainly focuses on two crucial components in the specific task: 1) proper generation of adaptive bins and 2) sufficient interaction between probability distribution and bins predictions. To specify, we employ the Transformer decoder to generate bins, novelly viewing it as a direct set-to-set prediction problem. We further integrate a multi-scale decoder structure to achieve a comprehensive understanding of spatial geometry information and estimate depth maps in a coarse-to-fine manner. Moreover, an extra scene understanding query is proposed to improve the estimation accuracy, which turns out that models can implicitly learn useful information from an auxiliary environment classification task. Extensive experiments on the KITTI, NYU, and SUN RGB-D datasets demonstrate that BinsFormer surpasses state-of-the-art monocular depth estimation methods with prominent margins. Code and pretrained models will be made publicly available.\footnote{\url{https://github.com/zhyever/Monocular-Depth-Estimation-Toolbox}}

\end{abstract}

\section{Introduction}
\begin{figure}[t]
    \begin{tabularx}{\textwidth}{@{}*{4}{C}c@{}}
    \rule{0pt}{13pt}\multirow{5}{*}{\includegraphics[width=1\linewidth]{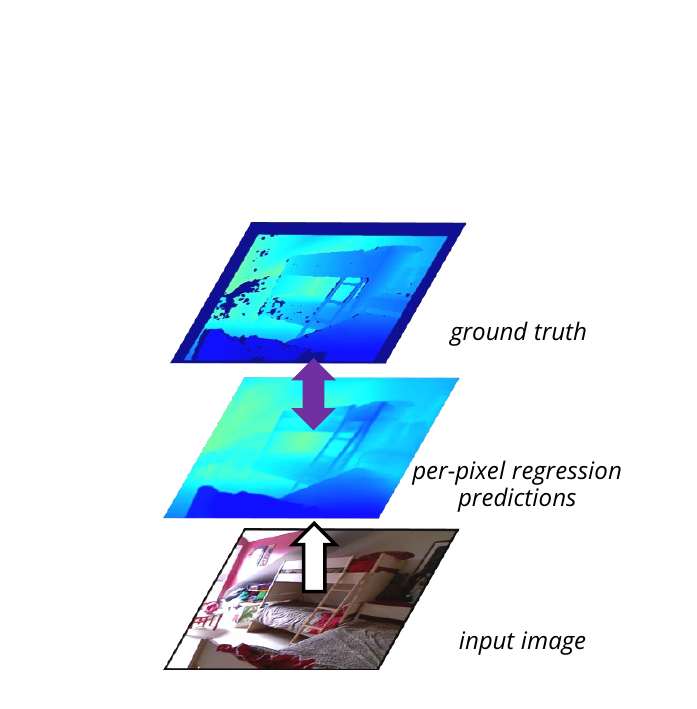}} & \rule{0pt}{20pt}\multirow{5}{*}{\includegraphics[width=1\linewidth]{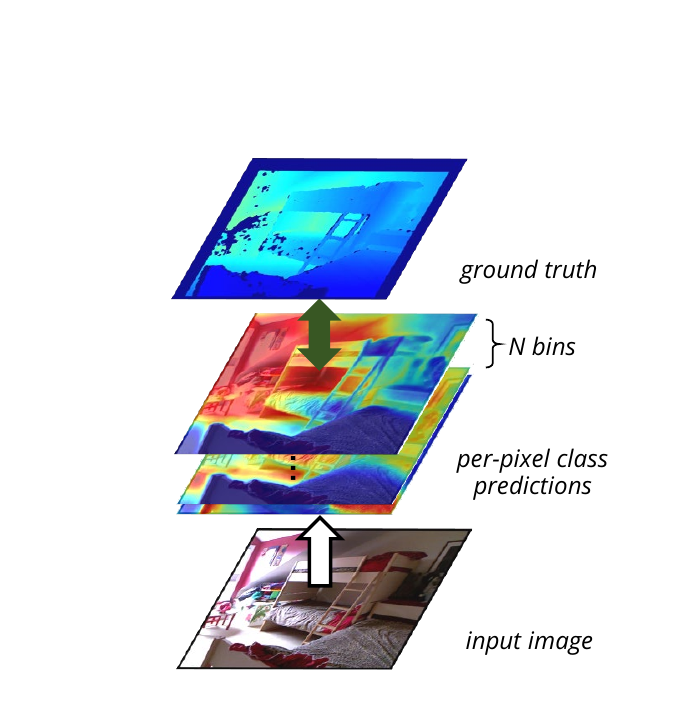}} & \rule{0pt}{13pt}\multirow{5}{*}{\includegraphics[width=1\linewidth]{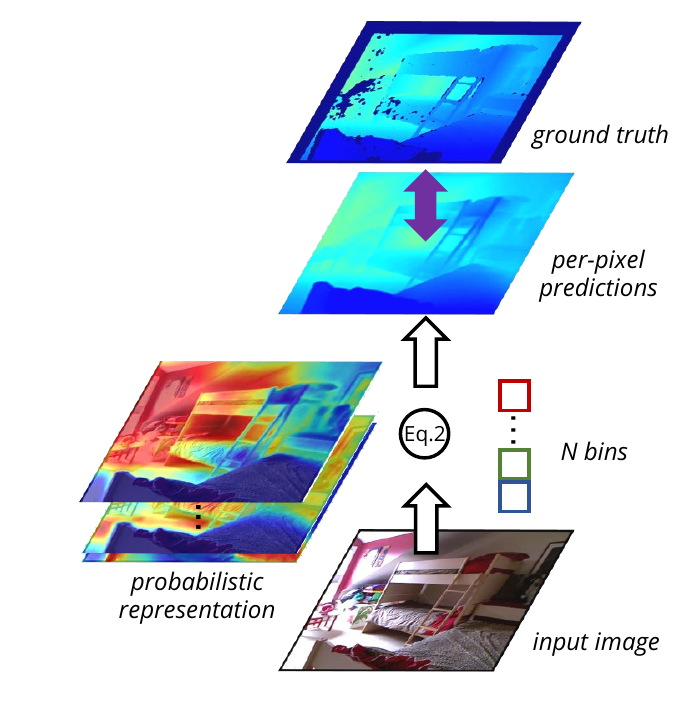}} & \rule{0pt}{13pt}\multirow{4}{*}{\includegraphics[width=1\linewidth]{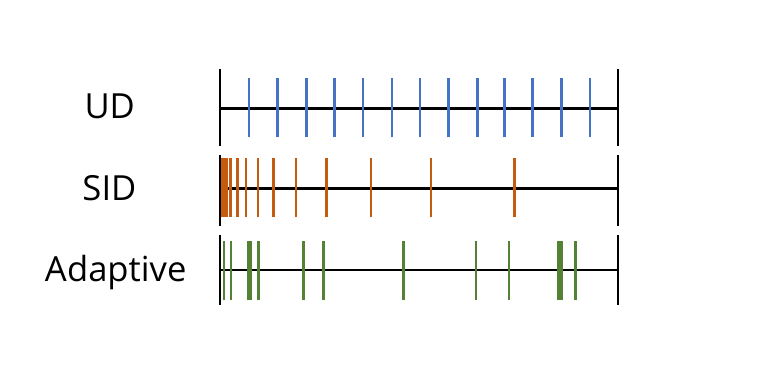}} \\
    \rule{0pt}{30pt}           &
    \rule{0pt}{30pt}           &
    \rule{0pt}{30pt}           &
    \rule{0pt}{32pt} \scriptsize {(d) bins types}        \\
    &&&     \\
    &&&  \multirow{2}{*}{\includegraphics[width=1.2\linewidth]{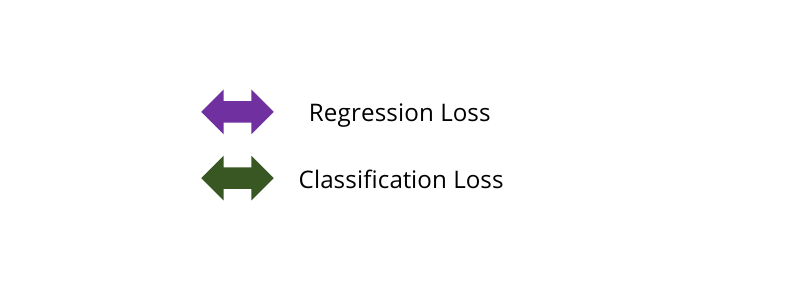}}     \\
    &&&     \\
    \scriptsize{(a) regression}    ~~~         &
    \scriptsize{(b) classification}  ~~~       &
    \scriptsize{(c) classification-reg.} &    
    \end{tabularx}
    \caption{\textbf{Regression} \textbf{\textit{vs.}} \textbf{classification} \textbf{\textit{vs.}} \textbf{classification-regression}. (a) Depth estimation with per-pixel regression directly predicts continuous depth and applies regression loss. (b) Classification predicts probabilistic representations and is supervised by classification loss. Predictions are discrete. (c) Classification-regression combines the best of both, which first predicts probabilistic representations and then gets continuous predictions via a linear combination with bins centers. (d) Commonly used bins types. Adaptive means adaptively generating bins based on input images.}
    \label{fig::formulation}
\end{figure}

Monocular depth estimation is a fundamental yet challenging task in computer vision, which requires the algorithm to predict each pixel's depth within a single input RGB image. So far, there have been numerous mainstream methods formulating depth estimation as per-pixel regression, such as DAV~\cite{huynh2020dav}, DPT~\cite{ranftl2021dpt} and TransDepth~\cite{yang2021transdepth} (Fig.~\ref{fig::formulation}a), where a regression loss is applied to each pixel prediction. Per-pixel regression methods can neatly predict pixel-wise depth, thus becoming a universal paradigm. Despite the proof of their great success, such methods still face problems of slow convergence and unsatisfied results~\cite{fu2018dorn}.

Another line of research~\cite{fu2018dorn,diaz2019soft} proposes to discretize continuous depth into several intervals and cast the depth network learning as a per-pixel \textit{classification} problem (Fig.~\ref{fig::formulation}b). While this strategy significantly improves the model performance, it is worth noting that the discretization of depth values will result in poor visual quality with apparent sharp discontinuities. 

To solve the issue, some methods~\cite{bhat2021adabins,johnston2020self} reformulate depth estimation as a per-pixel \textit{classification-regression} task (Fig.~\ref{fig::formulation}c), learning probabilistic representations on each pixel and predicting the final depth value as a linear combination with bin centers. The bin centers are pre-defined in Uniform/Log-uniform space (UD/SID) or trained ones (for each dataset). They combine the best of both tasks and achieve an improving performance. On top of that, Adabins~\cite{bhat2021adabins} observes the extreme variation of depth distribution among changing scenes and further proposes the adaptive bins generation module to predict bins centers adaptively. While Adabins~\cite{bhat2021adabins} boosts depth estimation performance to a remarkable extent, several dilemmas still exist. 1) It directly applies bins prediction depending on the highest resolution feature map (the output of the last layer of the Decoder), leading to the difficulty of squaring up the global information and perceiving scenes. While it utilizes a Transformer block, the facade encoder-decoder with limited receptive fields leads to an inevitable loss of global information. 2) Bins and probabilistic representations are predicted based on the same single layer feature, which lacks sufficient interactions and leads to the global and fine-grained information blemishing each other. 3) The proposed chamfer loss applying distribution constraints on bins estimation introduces futile inductive bias. All these problems impede models from predicting more accurate estimation results.

In this paper, we propose a conceptually simple yet effective approach called \textbf{BinsFormer}, tailored for classification-regression-based monocular depth estimation. It novelly views adaptive bins generation as a direct set prediction problem~\cite{carion2020detr}. We employ a separate Transformer decoder~\cite{dosovitskiy2020vit} to compute a set of pairs, each consisting of a bins length and a bins embedding vector. The bins embedding vector is used to get the probabilistic representations via a dot product with the per-pixel representations obtained from an underlying fully-convolutional decoder. Finally, BinsFormer predicts depth values as a linear combination of bins centers and probabilistic representations. Such disentangled decoder avoids bins embeddings and fine-grained per-pixel representations blemishing each other and combines the best of global and pixel-wise information. We further integrate a multi-scale decoder structure to comprehensively understand spatial geometry information and estimate depth maps in a coarse-to-fine manner. It enables sufficient interactions and aggregations of features via successive alternating cross-attention and self-attention mechanisms. To further improve the estimation accuracy, we equip an extra scene understanding query to the Transformer decoder, which aims to predict the classification of the input environment. It can benefit models to generate appropriate bins via auxiliary and implicit supervision.

We evaluate BinsFormer on three depth estimation datasets with various settings: NYU~\cite{silberman2012nyu} (indoor, depth range 0-10$m$), KITTI~\cite{geiger2013kitti} (outdoor, depth range 0-80$m$), and SUN-RGBD~\cite{song2015sunrgbd} (indoor, depth range 0-10$m$, directly fine-tuning). Numerous experiments domestrate BinsFormer achieves the new state-of-the-art on all these datasets with Swin-Transformer~\cite{liu2021swin} backbone, outperforming other methods with large margins. Exhaustive ablation studies further validate the effectiveness of each proposed component. % 2 pages + abs
\section{Related Work} % 1 page

\newcommand\etal{\textit{et al.}}

Monocular depth estimation plays a critical role in three-dimensional reconstruction and perception. There has been tremendous witnessed progress achieved by learning-based depth estimation methods in recent years. Eigen~\etal~\cite{eigen2014depth} groundbreakingly proposes a multi-scale deep network, consisting of a global network and a local network to predict the coarse depth and refine predictions, respectively. Motivated by~\cite{eigen2014depth}, convolutional architectures have been intensively studied for depth estimation. For instance, CLIFFNet~\cite{wang2020cliffnet} applies a multi-scale fusion convolutional framework to generate high-quality depth prediction. Recently, Transformer networks are gaining greater interest in the computer vision community~\cite{dosovitskiy2020vit,liu2021swin,carion2020detr,zheng2021setr}. Following the success of recent trends that apply the Transformer to solve computer vision tasks, TransDepth~\cite{yang2021transdepth} and DPT~\cite{ranftl2021dpt} propose to replace convolution operations with Transformer layers, which further boosts model performance.

Though the above methods have significantly improved depth prediction accuracy, they suffer from relatively slow convergence and sub-optimal solutions since they regard monocular depth estimation as a \textit{regression} task~\cite{fu2018dorn}. Another line of research~\cite{fu2018dorn,diaz2019soft} proposes to discretize continuous depth into several intervals and cast the depth network learning as a per-pixel \textit{classification} problem. DORN~\cite{fu2018dorn} designs an effective ordinal classification depth estimation loss and develops an ASPP~\cite{chen2017deeplab} module to extract multi-level information. Based on \cite{fu2018dorn}, \cite{diaz2019soft} softens the classification target during training and achieves improving performance. Moreover, some methods~\cite{bhat2021adabins,johnston2020self} reformulate the problem as \textit{classification-regression} to alleviate poor visual quality with apparent sharp depth discontinuities caused by discretization of depth values. Johnston~\etal~\cite{johnston2020self} introduces the stereo DDV for monocular depth estimation. To further improve the model performance, Adabins~\cite{bhat2021adabins} proposes an adaptive bins strategy, which is crucial for accurate depth estimation.

In this paper, we further investigate the adaptive bins strategy and propose \textbf{BinsFormer}, tailored for classification-regression-based monocular depth estimation. We novelly treat the adaptive bins generation as a set prediction process~\cite{carion2020detr} and develop Transformer layers to resort to the problem, which is intuitively different from previous work~\cite{ranftl2021dpt,yang2021transdepth} that only utilize Transformer to strengthen the encoder capability. Furthermore, we melt an effective multi-scale refinement strategy into the Transformer-based decoder in a neat fashion. An extra scene understanding task further improves the model performance.

Indeed, various strategies have been explored to benefit monocular depth estimation, such as self-supervised learning~\cite{godard2019digging}, multi-task training~\cite{jung2021fine}, specific supervision losses~\cite{wang2020cliffnet}, sparse ordinal~\cite{lo2021depth} or relative depth estimation~\cite{lee2019monocular}. Our method focuses on the most basic framework design, which can be plugged into any other method as a strong baseline.
 % 1 page / half
\section{Methods} % 1 page

In this section, we first present the overview of BinsFormer. Then, we introduce our instantiation of the adaptive bins generation strategy with the help of Transformer decoder~\cite{carion2020detr}. Finally, we present the auxiliary scene understanding task and the multi-scale prediction refinement strategies, which can be neatly melted into the framework and improve the depth estimation performance.

\subsection{Framework Overview}

\begin{figure}[t]
    \includegraphics[width=1\linewidth]{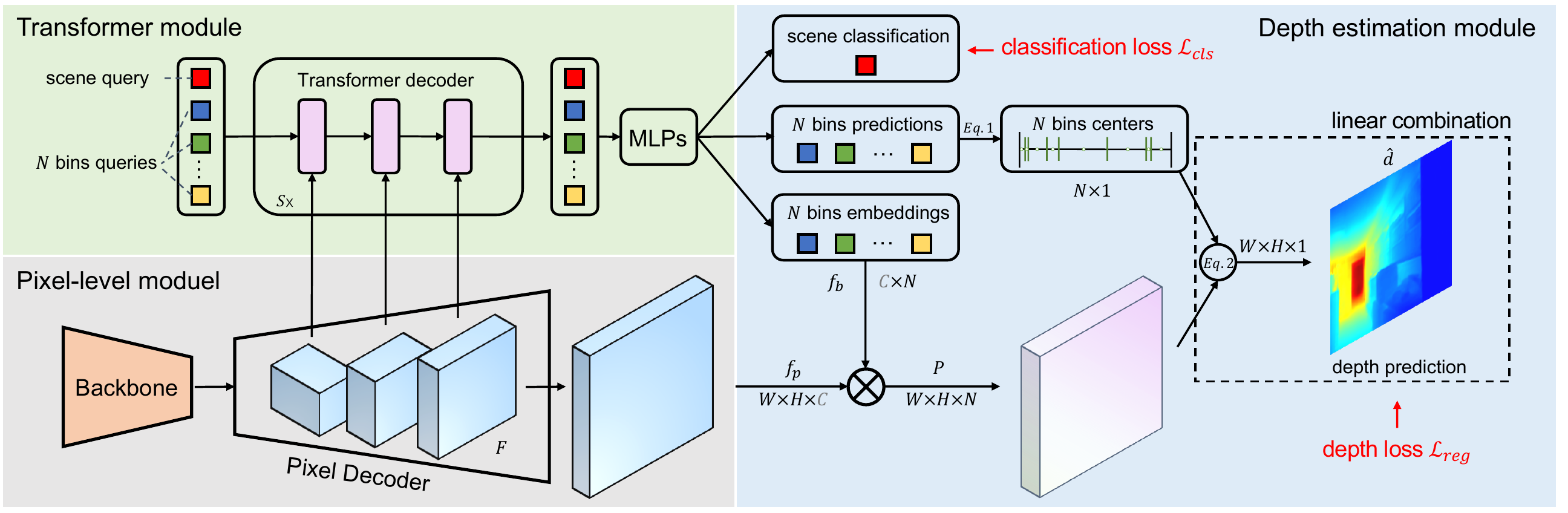}
    \caption{BinsFormer overview: We use a backbone and a pixel decoder to extract and upsample image features. A transformer decoder attends to multi-scale image features $\textbf{F}$ and generate $1+N$ output embedding. The first one predicts enviroment classification and the other $N$ ones independently predict $N$ bins lengths $\textbf{b}$ and $N$ bins embeddings $f_b$, respectively. Then the model predicts the probability distribution map $P$ via a dot product between pixel representations $f_p$ and bins embeddings $f_b$. Note, the dimensions for the dot $\bigotimes$ are shown in gray. The final depth estimation is calculated by a linear combination between the probability distribution map $P$ and post-processed $N$ bins centers $c(\textbf{b})$.}
    \label{fig::framework}
\end{figure}

BinsFormer mainly consists of three essential components (see Fig.~\ref{fig::framework}): the pixel-level module, the Transformer module, and the depth estimation module. Moreover, we propose the auxiliary scene classification and the multi-scale prediction refinement strategies to further boost model performance.

Given an input RGB image, the pixel-level module first extracts image features and decodes them into multi-scale immediate features $\mathbf{F}$ and the per-pixel representations $f_p$. Benefiting from the encoder-decoder framework with skip connections, BinsFormer can fully extract local information for fine-grained depth estimation. Then, queries in the Transformer module interact with $\mathbf{F}$ with the help of attention mechanisms. Independent MLPs are applied to project the query embeddings into bins predictions $\textbf{b}$ and bins embeddings $f_b$. Novelly viewing the bins generation as a set-to-set prediction problem and applying Transformer, BinsFormer can explore the global information and predict appropriate bins for integral depth estimation. Finally, the depth estimation module aggregates the best of the abovementioned modules and predicts final depth. It first calculates the probability distributions $P$ and then combines them with bins centers $c(\textbf{b})$ via linear combinations.

On top of that, we equip an extra scene understanding query to the Transformer decoder, which aims to predict the classification of the input environment. Similarly, an MLP projects the query embedding to the classification result. The extra task can benefit models to generate appropriate bins via auxiliary and implicit supervision. Moreover, we further integrate a multi-scale decoder structure to comprehensively understand spatial geometry information and estimate depth maps in a coarse-to-fine manner. The Transformer queries progressively interact with multi-scale features $\mathbf{F}$, enabling sufficient aggregations of features via successive attention modules.

\subsection{BinsFormer}

\subsubsection{Per-pixel module} takes an image of size $H\times W$ as input. A backbone is applied to extract a set of feature maps. Then, a commonly used decoder gradually upsamples the features to generate the per-pixel representation $f_{p} \in \mathbb{R}^{H\times W\times C}$, where $C$ is the representation dimension. It produces $S$ scale immediate feature maps ${\mathbf{F}}=\{f_i\}_{i=1}^S$ as well, where $f_i$ indicates the feature map at scale $i^{th}$. 

Note that any per-pixel depth estimation model fits the pixel-level module design including Transformer-based models~\cite{ranftl2021dpt,yang2021transdepth,bhat2021adabins}. However, unlike previous methods~\cite{ranftl2021dpt,yang2021transdepth} that only adopt Transformer to replace convolutional operations in models, BinsFormer seamlessly converts Transformer to solve the bins generation and leaves the per-pixel backbone untouched.

\subsubsection{Transformer module} applies the standard Transformer decoder~\cite{dosovitskiy2020vit}, transforming $N$ embeddings using multi-head self- and cross-attention mechanisms. Following~\cite{carion2020detr}, BinsFormer decodes the $N$ bins in parallel at each decoder layer. The $N$ input embeddings serve as \textit{bins queries} to interact with image features $\textbf{F}$ and are transformed into an output embedding $ Q\in \mathbb{R}^{C_{q}\times N}$ by the decoder. Then, we apply a linear perceptron with softmax on top of the output embeddings $Q$ to yield $N$ bins length $\textbf{b} = \{b_i\}_{i=1}^N$. Moreover, we utilize a 3-layer perceptron with ReLU activation function on $Q$ to predict $N$ bins embeddings $f_b\in \mathbb{R}^{C\times N}$ to calculate the similarity with per-pixel representations $f_p$ in the depth prediction module.

\subsubsection{Depth prediction module} aggregates outputs from the per-pixel module and the Transformer module to predict depth. Given the predicted bins lengths $\textbf{b}$ from the Transformer module, it first converts them to bins centers via a simple post-process following \cite{bhat2021adabins}:
\begin{equation}
    c(b_i) = d_{min} + \left(d_{max}-d_{min}\right)\left(\frac{b_i}{2}+\sum\limits_{j=1}^{i-1}b_j\right),
\end{equation}
\noindent where $c(b_i)$ is center depth of the $i^{th}$ bins. $d_{max}$ and $d_{min}$ are the max and the min valid depth values of the dataset, respectively.

Meanwhile, we obtain a similarity map via a dot product between the pixel-wise representations $f_p$ from the per-pixel module and the bins embeddings $f_b$ from the Transformer module. We then convert it to a probability distribution map $P\in \mathbb{R}^{H\times W\times N}$ by a \texttt{Softmax} fuction. Finally, at each pixel, the final depth value $\hat d$ is calculated from a linear combination of the probability distribution at that pixel and the depth-bin-centers $c(\textbf{b})$ as follows:
\begin{equation}
    \hat d = \sum\limits_{i=1}^{N}c(b_i)p_i.
\end{equation}

Compared to Adabins~\cite{bhat2021adabins}, we disentangle the bins generation and avoid bins embeddings and fine-grained per-pixel representations blemishing each other. This enables us to predict more accurate depth without large-area failures, as can be seen in Fig.~\ref{fig::figure-nyu}.

After predicting final depth maps, we apply a scaled version of the Scale-Invariant loss (SI) introduced by Eigen \etal~\cite{eigen2014depth}:
\begin{equation}
    \mathcal{L}_{reg} = \alpha \sqrt{\frac{1}{T}\sum_i g_{i}^{2} - \frac{\lambda}{T^2}(\sum_i g_i)^2},
    \label{eq::loss}
\end{equation}
\noindent where $g_i = \log \Tilde{d_i} - \log d_i$ with the ground truth depth $d_i$ and predicted depth $\Tilde{d_i}$. $T$ denotes the number of pixels having valid ground truth values. Following~\cite{bhat2021adabins}, we use $\lambda = 0.85$ and $\alpha = 10$ for all our experiments.

\begin{figure}[t]
    \includegraphics[width=1\linewidth]{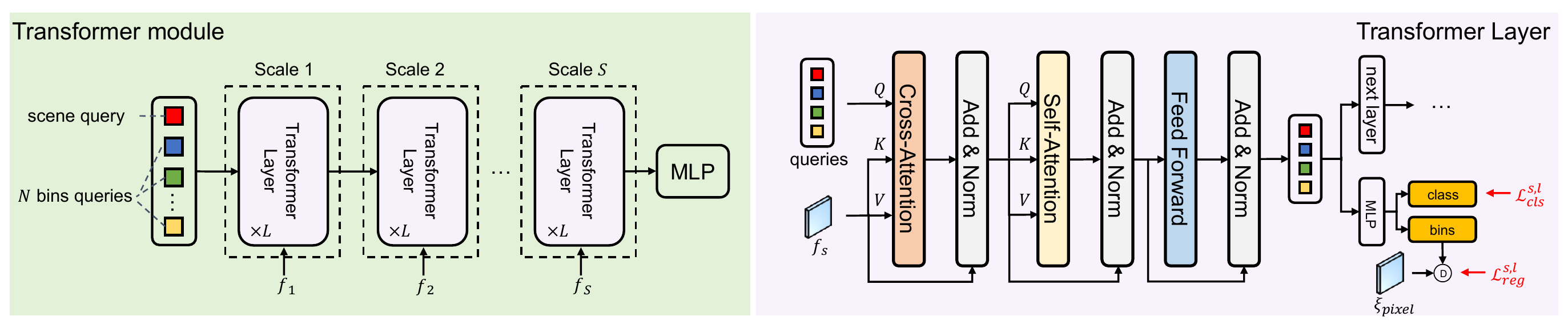}
    \caption{Multi-scale prediction refinement: we propose to apply progressive interaction and aggregation among the embeddings and image features. At each scale of the Transformer decoder, one resolution of the multi-scale feature is fed into the Transformer layer. Multi-scale losses with gradually strengthened punishment weights are applied on each Transformer layer.}
    \label{fig::multi-scale}
\end{figure}

\subsubsection{Auxiliary scene classification} is an auxiliary subtask to provide implicit guidance to the bins generation. Beyond the bins queries, we equip the Transformer decoder with an extra \textit{scene query}, which is used to classify the scene environment. It can learn the global semantic information and transfer such knowledge through successive alternating self-attention to the bins queries. Similar to the bins embeddings, we adopt a 3-layer perceptron with ReLU activation function on the output embedding to yield the final classification result. During training, a simple \texttt{CrossEntropy} classification loss $\mathcal{L}_{cls}$ is applied.

Unlike Adabins~\cite{bhat2021adabins}, which applies chamfer loss to constrain the distribution of bins, our method avoids introducing such futile inductive bias. It means the supervision from the auxiliary classification subtask is implicit. The bins queries can adaptively absorb the global semantic information by self-attention with the scene query. This strategy only leads to a negligible overhead during the training process, compared with the computational pixel-wise chamfer loss.

\subsubsection{Multi-scale prediction refinement} is applied to obtain a global understanding of the image structure information and exploit a coarse-to-fine depth refinement. It is intuitively reasonable that the model can effectively square up the global information in low-resolution feature maps where the structure information is well reserved while high-frequency details are discarded. Similarly, since we combine the bins and the per-pixel representations to predict final depth, learning the fine-grained details in high-resolution feature maps is crucial for high-quality depth estimation. Hence, we propose the multi-scale prediction refinement strategy, which can be seamlessly adopted with the help of the Transformer module.

As shown in Fig~\ref{fig::multi-scale}, we feed one resolution of the multi-scale feature $\textbf{F}$ to one Transformer decoder layer at a time. Each scale of the Transformer decoder contains $L$ Transformer layers that consists of a cross-attention module, a self-attention module and a feed-forward network (FFN). During the forward propagation, queries at scale $s$ first query the input image feature $f_s$ via a cross-attention module, and then aggregate information among themselves through a self-attention module. Following the standard Transformer~\cite{vaswani2017attention}, a FFN is applied to improve the model capacity. Queries are zero initialized before being fed into the Transformer decoder and are associated with learnable positional embeddings. We also apply the depth estimation module at each Transformer layer to provide auxiliary supervison. Therefore, the total loss can be formulated as:
\begin{equation}
    \mathcal{L}_{total} = \sum\limits_{s=1}^{S}\left(
        w_s\sum\limits_{l=1}^{L}\left(
            \mathcal{L}_{reg}^{s,l} + \mu\mathcal{L}_{cls}^{s,l}
        \right)
    \right),
    \label{eq::total-loss}
\end{equation}
\noindent where $\mu=1e^{-3}$ is a hpyerparameter to balance the auxiliary classification loss and the depth estimation loss. $w$ is a scale weight, which is set to $\{\frac{1}{2^s}, \frac{1}{2^{s-1}}, \cdots, 1\}$. By simply reweight the multi-scale punishment to estimation results, we seamlessly integrate the multi-scale refinement strategy to the Transformer module.

 % 5 pages
\section{Experiments}
\label{sec:experiments}
In this section, we evaluate the performance of BinsFormer by comparing it against several baselines, starting by introducing datasets, evaluation metrics, and implementation details. Then, we present the comparison to the state-of-the-art methods (containing a cross-dataset generalization evaluation), ablation studies, and uncertainty predictions.

\subsection{Datasets and evaluation metrics}

\subsubsection{Datasets:} We assess the proposed method using KITTI~\cite{geiger2013kitti}, NYU-Depth-v2~\cite{silberman2012nyu}, and SUN RGB-D~\cite{song2015sun} datasets. \textbf{KITTI} is a dataset that provides stereo images and corresponding 3D laser scans of outdoor scenes captured by equipment mounted on a moving vehicle~\cite{geiger2013kitti}. Following the standard Eigen training/testing split~\cite{eigen2014depth}, we use around 26K images from the left view for training and 697 frames for testing. When evaluation, we use the crop as defined by Garg et al.~\cite{garg2016unsupervised} and upsample the prediction to the ground truth resolution. For the online KITTI depth prediction, we use the official benchmark split~\cite{uhrig2017sparsity}, which contains around 72K training data, 6K selected validation data, and 500 test data without the ground truth.  \textbf{NYU-Depth-v2} provides images and depth maps for different indoor scenes. We train our network on a 50K RGB-Depth pairs subset following previous works. We evaluate the results on the pre-defined center cropping by Eigen et al.~\cite{eigen2014depth}. \textbf{SUN RGB-D} is an indoor dataset consisting of around 10K images with high scene diversity collected with four different sensors~\cite{song2015sun}. We apply this dataset for generalization evaluation. Specifically, we cross-evaluate our NYU pre-trained models on the official test set of 5050 images without further fine-tuning. The depth upper bound is set to 10 meters. Notably, this dataset is only for evaluation. We do not train on this dataset.

\subsubsection{Evaluation metrics:} We follow the standard evaluation protocol of the prior work~\cite{eigen2014depth} to confirm the effectiveness of BinsFormer in experiments. For the NYU Depth V2, the KITTI Eigen split datasets, and the SUN RGB-D dataset, we utilize the accuracy under the threshold ($\delta_i < 1.25^i, i = 1, 2, 3$), mean absolute relative error (AbsRel), mean squared relative error (SqRel), root mean squared error (RMSE), root mean squared log error (RMSElog) and mean log10 error (log10) to evaluate our methods. In terms of the online KITTI depth prediction benchmark~\cite{uhrig2017sparsity}, we use the scale-invariant logarithmic error (SILog), percentage of AbsRel and SqRel (absErrorRel, sqErrorRel), and root mean squared error of the inverse depth (iRMSE).

\subsection{Implementation details}
\subsubsection{Training settings:} BinsFormer is implemented with the Pytorch~\cite{paszke2019pytorch} framework. We train the entire model with the batch size 16, learning rate $1e-4$ for 38.4k iterations on a single node with 8 NVIDIA V100 32GB GPUs, which takes around 5 hours. We utilize the AdamW optimizer~\cite{kingma2014adam} with ($\beta_1$, $\beta_2$, $wd$) = (0.9, 0.999, 0.01), where $wd$ is the weight decay. The linear learning rate warm-up strategy is applied for the first 30\% iterations. The cosine annealing learning rate strategy is adopted for the learning rate decay. For the NYU-Depth-v2 dataset, we utilize the official 25 classes divided by folder names for the auxiliary scene understanding task. For KITTI, since the outdoor dataset is tough to classify, we omit the scene classification loss and only use ground truth depth to provide supervision.

\subsubsection{Backbone:} BinsFormer is compatible with any backbone architecture. In our work we use the standard convolution-based ResNet~\cite{he2016resenet} backbones (ResNet-18 and ResNet-50, respectively) and recently proposed Transformer-based Swin-Transformer~\cite{liu2021swin} backbones.

\subsubsection{Pixel decoder:} As for the pixel decoder in Fig.~\ref{fig::framework}, any depth estimation decoder can be adopted (\textit{e.g.,}~\cite{fu2018dorn,lee2019bts,alhashim2018densedepth}). There are numerous depth estimation methods use modules like ASPP~\cite{chen2017deeplab} or CBAM~\cite{woo2018cbam} to capture long-range correspondings. Since our Transformer module attends to all image representations, collecting both the global and local information to generate bins, we can omit the computional context aggregation in per-pixel module. Therefore, following~\cite{cheng2021maskformer}, a light-weight pixel decoder is applied based on the popular FPN network~\cite{lin2017fpn}.   

\subsubsection{Transformer decoder:} We stack $L=3$ Transformer layers for each scale of prediction refinement (\textit{i.e.,} 9 layers total) and 64 queries by default. The auxiliary loss is added to every intermediate Transformer decoder layer. Following~\cite{cheng2021mask2former}, we adopt a simple deformable encoder~\cite{zhu2020deformabledetr} to enhance the multi-scale image features. In our experiments, we observe that BinsFormer is competitive for depth estimation with a single decoder layer as well.

\begin{table*}[t]
    \centering
    \caption{Comparison of performances on the KITTI dataset. The reported numbers are from the corresponding original papers. Measurements are made for the depth range from $0m$ to $80m$. Best / Second best results are marked \textbf{bold} / \underline{underlined}. E-B5 are short for EfficientNet-B5~\cite{tan2019efficientnet}. \dag and \ddag represent the models are pre-trained by ImageNet-22K and auxiliary depth estimation dataset, respectively.}
    \label{tab::results-kitti-val}
    \scalebox{0.7}{%
        \begin{tabular}{|c|c|c|c|c|c|c|c|c|c|}
            \hline
            Method & Ref & Backbone & ~~~\textbf{$\delta_1$}$\uparrow$~~~ & ~~~\textbf{$\delta_2$}$\uparrow$~~~ & ~~~\textbf{$\delta_3$}$\uparrow$~~~ & ~REL~$\downarrow$~ & Sq-rel~$\downarrow$ & RMS~$\downarrow$ & RMS log~$\downarrow$ \\
            \hline
            % Saxena~\etal~\cite{saxena2005learning} & - & 0.601 & 0.820  & 0.926 & 0.280   & 3.012  & 8.734 & 0.361    \\
            % Eigen~\etal~\cite{eigen2014depth} & Convolutions & 0.702 & 0.898 & 0.967 & 0.203   & 1.548  & 6.307 & 0.282    \\
            % Liu~\etal~\cite{liu2015learning} & Convolutions & 0.680 & 0.898 & 0.967 & 0.201   & 1.584  & 6.471 & 0.273    \\
            Godard~\etal~\cite{godard2017unsupervised}& CVPR 2017 & ResNet-50 & 0.861 & 0.949 & 0.976 & 0.114   & 0.898  & 4.935 & 0.206 \\
            Johnston~\etal~\cite{johnston2020self}& CVPR 2020 & ResNet-101 & 0.889 & 0.962 & 0.982 & 0.106 & 0.861 & 4.699 & 0.185 \\
            Gan~\etal~\cite{gan2018monocular}& ECCV 2018 & ResNet-101 & 0.890 & 0.964 & 0.985 & 0.098   & 0.666  & 3.933 & 0.173 \\
            % PLADE-Net~\cite{gonzalez2021plade} & FAL-Net & 0.900 & 0.967 & 0.985 & 0.089 & 0.590 & 4.008 & 0.172\\
            DORN~\etal~\cite{fu2018dorn}& CVPR 2018 & ResNet-101 & 0.932 & 0.984 & 0.994 & 0.072   & 0.307  & 2.727 & 0.120\\
            Yin~\etal~\cite{yin2019enforcing}& ICCV 2019 & ResNext-101 & 0.938 & 0.990  & \underline{0.998} & 0.072 & -- & 3.258 & 0.117 \\
            PGA-Net~\cite{xu2020probabilistic}& TPAMI 2020 & ResNet-50 & 0.952 & 0.992 & \underline{0.998} & 0.063 & 0.267 & 2.634 & 0.101 \\
            BTS~\cite{lee2019bts}& Arxiv 2019 & DenseNet-161& 0.956 & 0.993 & \underline{0.998} & 0.059   & 0.245  & 2.756 & 0.096 \\
            TransDepth~\cite{yang2021transdepth}& ICCV 2021 & ResNet-50+ViT-B\dag & 0.956 & 0.994 & \textbf{0.999} & 0.064 & 0.252  & 2.755 & 0.098 \\
            DPT~\cite{ranftl2021dpt} & ICCV 2021 &ResNet-50+ViT-B\ddag & 0.959 & 0.995 & \textbf{0.999} & 0.062 & --& 2.573 & 0.092 \\
            AdaBins~\cite{bhat2021adabins} & CVPR 2021 &E-B5+mini-ViT & 0.964 & 0.995 & \textbf{0.999} & 0.058  & 0.190  & 2.360 & 0.088\\
            \hline
            \multirow{6}{*}{\textbf{BinsFormer}} & \multirow{6}{*}{\textbf{Ours}}
             & ResNet-18  & 0.954 & 0.994 & \textbf{0.999} & 0.065 & 0.230 & 2.574 & 0.099 \\
             && ResNet-50 & 0.962 & 0.994 & \textbf{0.999} & 0.061 & 0.208 & 2.426 & 0.093 \\
             && Swin-Tiny & 0.968 & 0.995 & \textbf{0.999} & 0.058 & 0.183 & 2.286 & 0.088 \\
             && Swin-Base & \underline{0.970} & \underline{0.996} & \textbf{0.999} & 0.056 & 0.172 & 2.248 & 0.085 \\
             && Swin-Base\dag & \textbf{0.974} & \textbf{0.997} & \textbf{0.999} & \underline{0.053} & \underline{0.156} & \underline{2.141} & \underline{0.080} \\
             && Swin-Large\dag & \textbf{0.974} &  \textbf{0.997} & \textbf{0.999} & \textbf{0.052}& \textbf{0.151} & \textbf{2.098} & \textbf{0.079} \\
            \hline
        \end{tabular}
    }
\end{table*} 
\begin{table}[t]
    \centering
    \caption{Comparison of performances on the KITTI depth estimation benchmark test set. Reported numbers are from the official benchmark website.}
    \scalebox{0.88}{%
        \begin{tabular}{@{}|l|c|c|c|c|@{}}
            \hline
            Method  & ~~~~SILog$\downarrow$~~~~ & sqErrorRel$\downarrow$ & absErrorRel$\downarrow$  & ~~~iRMSE$\downarrow$~~~ \\ \hline
            DORN~\cite{fu2018dorn} & 11.77  & 2.23 & 8.78 & 12.98\\
            BTS~\cite{lee2019bts}  & 11.67  & 2.21 & 9.04 & 12.23     \\
            BANet~\cite{aich2020BANet} & 11.55  & 2.31 & 9.34 & 12.17     \\
            PWA~\cite{lee2021PWA} & 11.45  & 2.30 & 9.05 & 12.32      \\
            ViP-DeepLab~\cite{qiao2021vip} & \underline{10.80}  & \underline{2.19} & \underline{8.94} & \underline{11.77}  \\ 
            \hline
            \textbf{BinsFormer}  & \textbf{10.14}  & \textbf{1.69} & \textbf{8.23} & \textbf{10.90} \\
            \hline
        \end{tabular}
    }
    
\label{tab::results-kitti-test}
\vspace{-0.2cm}
\end{table}

\subsection{Comparison with the state-of-the-art}
This section compares the proposed approach with the current state-of-the-art monocular depth estimation methods.

\subsubsection{KITTI}: We evaluate on the Eigen split~\cite{eigen2014depth} and Tab.~\ref{tab::results-kitti-val} reports the results. BinsFormer significantly outperforms all the leading methods. Qualitative comparisons can be seen in the \textit{supplementary material}. We then evaluate the proposed method on the online KITTI depth prediction benchmark server~\footnote{\url{http://www.cvlibs.net/datasets/kitti/eval\_depth.php?benchmark=depth\_prediction}} and report the results in Tab.~\ref{tab::results-kitti-test}. While a saturation phenomenon persists in SILog, BinsFormer still achieves 6.1\% improvement on this metric.

\subsubsection{NYU-Depth-v2}
\begin{figure}[b!]
    \centering
    \footnotesize
    \begin{tabular}{l}
        \includegraphics[width=0.95\linewidth]{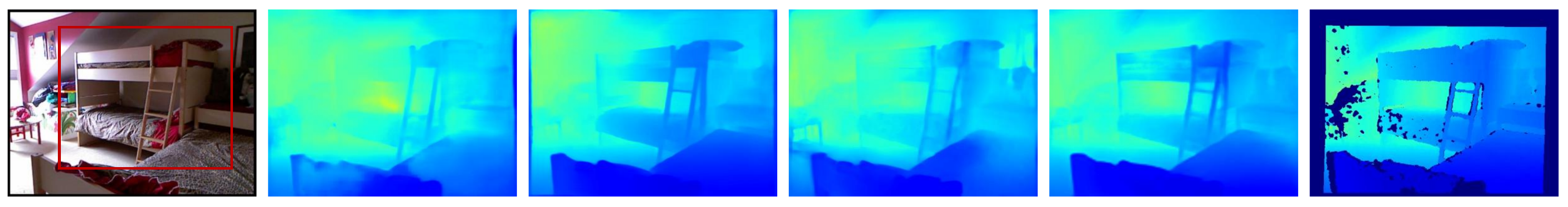} \vspace{-0.2cm}\\
        \includegraphics[width=0.95\linewidth]{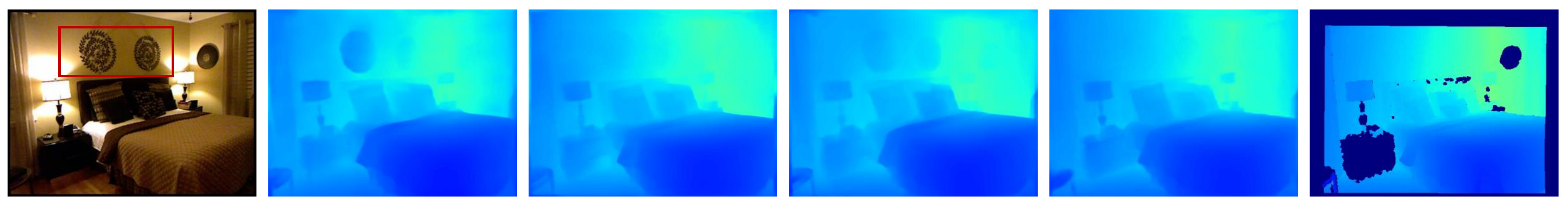} \vspace{-0.2cm}\\
        \includegraphics[width=0.95\linewidth]{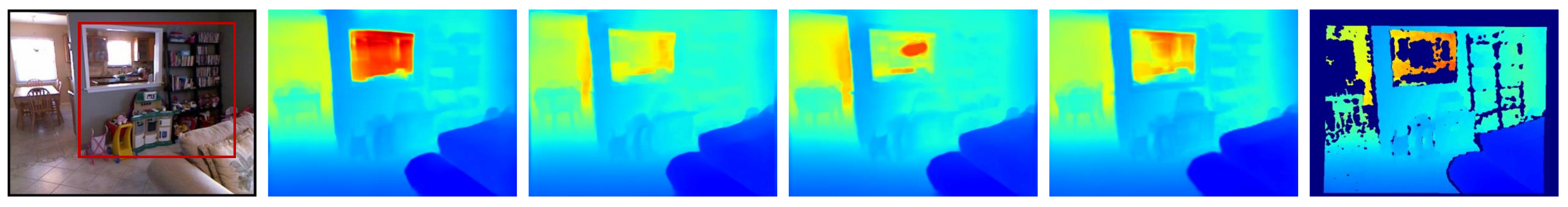} \vspace{-0.2cm}\\
        \includegraphics[width=0.95\linewidth]{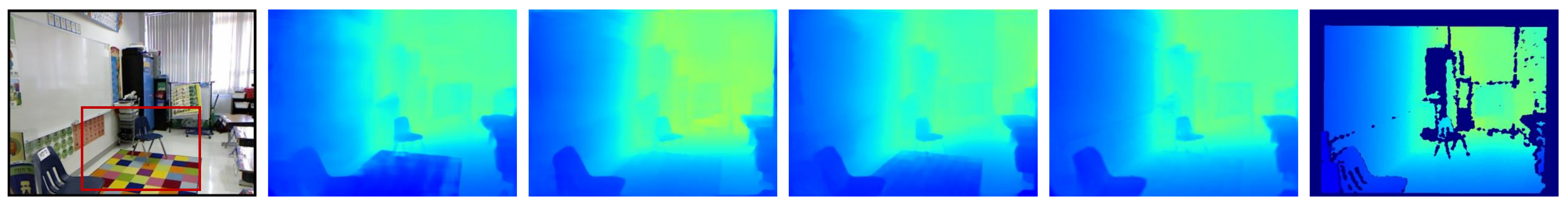} \vspace{-0.2cm}\\
        \includegraphics[width=0.95\linewidth]{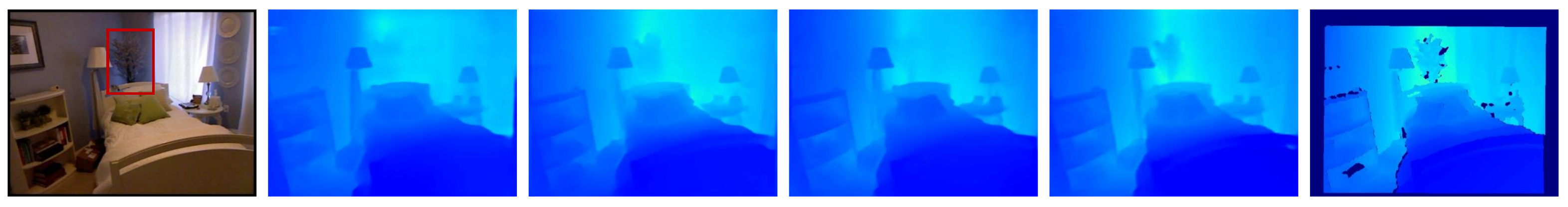} \\
        \hspace{0.055\linewidth}RGB
        \hspace{0.07\linewidth}BTS~\cite{lee2019bts}
        \hspace{0.04\linewidth}DPT~\cite{ranftl2021dpt}
        \hspace{0.03\linewidth}Adabins~\cite{bhat2021adabins}
        \hspace{0.05\linewidth}Ours
        \hspace{0.1\linewidth}GT\\
    \end{tabular}
    \caption{Qualitative comparison on the NYU-Depth-v2 dataset.}
    \label{fig::figure-nyu}
\end{figure}

: Tab.~\ref{tab::results-nyu} lists the performance comparison results on the NYU-Depth-v2 dataset. While the performance of the state-of-the-art models tends to approach saturation, BinsFormer outperforms all the competitors with prominent margins in all metrics. It indicates the effectiveness of our proposed methods. Qualitative comparisons can be seen in Fig.~\ref{fig::figure-nyu}.

\subsubsection{SUN RGB-D}: Following Adabins~\cite{bhat2021adabins}, we conduct a cross-dataset evaluation by training our models on the NYU-Depth-v2 dataset and evaluating them on the test set of the SUN RGB-D dataset without any fine-tuning. As shown in Tab.~\ref{tab::generalization}, significant improvements in all the metrics indicate an outstanding generalization performance of BinsFormer. Qualitative results are shown in the supplementary material.

\begin{table}[t!]
    \caption{Comparison of performances on the NYU-Depth-v2 dataset. The reported numbers are from the corresponding original papers.  We provide results of BinsFormer based on various encoder to demonstrate the superior performance.}
    \centering
    % \small
    \scalebox{0.89}{%
        \begin{tabular}{|c|c|c|c|c|c|c|c|c|}
            \hline
            \textbf{Method}    & ~~~~~\textbf{Encoder}~~~~~ & ~~~\boldsymbol{$\delta_1$}$\uparrow$~~ & ~~~\boldsymbol{$\delta_2$}$\uparrow$~~    & ~~~\boldsymbol{$\delta_3$}$\uparrow$~~     & ~\textbf{REL}$\downarrow$   & ~\textbf{RMS}$\downarrow$  & ~\boldsymbol{$log_{10}$}$\downarrow$ \\
            \hline
            DORN~\cite{fu2018dorn}& ResNet-101 &0.828 & 0.965 & 0.992 & 0.115 & 0.509 & 0.051\\
            Yin~\etal~\cite{yin2019enforcing} & ResNeXt-101 & 0.875 & 0.976 & 0.994 & 0.108 & 0.416 & 0.048 \\
            BTS~\cite{lee2019bts}& DenseNet-161 &0.885 & 0.978 & 0.994 & 0.110 & 0.392 & 0.047\\
            DAV~\cite{huynh2020dav}& DRN-D-22 &0.882 & 0.980 & 0.996 & 0.108 & 0.412 & --\\
            TransDepth~\cite{yang2021transdepth} & Res-50+ViT-B\dag & 0.900 & 0.983 & 0.996 & 0.106 & 0.365 & 0.045\\
            DPT~\cite{ranftl2021dpt} & Res-50+ViT-B\ddag & 0.904 & \underline{0.988} & \textbf{0.998} & 0.110 & 0.357 & 0.045 \\
            AdaBins~\cite{bhat2021adabins} & E-B5+Mini-ViT & 0.903 & 0.984 & 0.997 & 0.103 & 0.364 & 0.044 \\
            \hline
            \multirow{6}{*}{\textbf{BinsFormer}}
             & ResNet-18 & 0.866 & 0.976 & 0.994 & 0.122 & 0.410 & 0.050 \\
             & ResNet-50 & 0.882 & 0.978 & 0.995 & 0.111 & 0.390 & 0.047 \\
             & Swin-Tiny & 0.890 & 0.983 & 0.996 & 0.113 & 0.379 & 0.047 \\
             & Swin-Base & 0.902 & 0.984 & 0.996 & 0.104 & 0.362 & 0.044 \\
             & Swin-Base\dag & \underline{0.917} & \underline{0.988} & \underline{0.997} & \underline{0.098} & \underline{0.340} & \underline{0.041} \\
             & Swin-Large\dag & \textbf{0.925} & \textbf{0.989} & \underline{0.997} & \textbf{0.094} & \textbf{0.330} & \textbf{0.040} \\
            \hline
        \end{tabular} 
    }
    \label{tab::results-nyu}
\end{table}
\begin{table}[t!]
    \caption{Results of models trained on the NYU-Depth-v2 dataset and tested on the SUN RGB-D dataset \cite{song2015sun} without fine-tuning.}
    \centering
    % \small
    \scalebox{0.89}{%
        \begin{tabular}{|c|c|c|c|c|c|c|c|c|}
            \hline
            \textbf{Method}    & ~~~~~\textbf{Encoder}~~~~~ & ~~~\boldsymbol{$\delta_1$}$\uparrow$~~ & ~~~\boldsymbol{$\delta_2$}$\uparrow$~~    & ~~~\boldsymbol{$\delta_3$}$\uparrow$~~     & ~\textbf{REL}$\downarrow$   & ~\textbf{RMS}$\downarrow$  & ~\boldsymbol{$log_{10}$}$\downarrow$ \\
            \hline
            Chen~\etal~\cite{chen2019structure} & SENet~\cite{hu2018senet} & 0.757 & 0.943 & 0.984 & 0.166 & 0.494 & 0.071\\
            Yin~\etal~\cite{yin2019enforcing} & ResNeXt-101 & 0.696 & 0.912 & 0.973 & 0.183 & 0.541 & 0.082\\
            BTS~\cite{lee2019bts} & DenseNet-161 & 0.740 & 0.933 & 0.980 & 0.172 & 0.515 &  0.075 \\ 
            Adabins~\cite{bhat2021adabins} & E-B5+Mini-ViT & \underline{0.771} & 0.944 & 0.983 & \underline{0.159} & \underline{0.476} & \underline{0.068} \\ 
            \hline
            \multirow{3}{*}{\textbf{BinsFormer}} 
            & ResNet-18 & 0.738 & 0.935 & 0.982 & 0.175 & 0.504 & 0.074 \\
            & Swin-Tiny & 0.760 & \underline{0.945} & \underline{0.985} & 0.162 & 0.478 & 0.069 \\
            & Swin-Large\dag & \textbf{0.805} & \textbf{0.963} & \textbf{0.990} & \textbf{0.143} & \textbf{0.421} & \textbf{0.061} \\
            \hline
        \end{tabular} 
    }
    \label{tab::generalization}
\end{table}

\subsection{Ablation studies}

\begin{table}[b!]
    \caption{Ablation study results on the NYU dataset. We compare BinsFormer with regression based methods and  classification-regression based methods with different bins generalization strategies. Furthermore, we investigate the effectiveness of each component in BinsFormer.}
    \label{tab::ablation-nyu}
    \centering
    % \small
    \scalebox{0.78}{%
        \begin{tabular}{|c|c|c|c|c|c|c|c|c|}
        \hline
        \textbf{Method} & ~~\textbf{Cls.-Reg.}~~ & ~\textbf{Ada. Bins}~ & \textbf{Bins Query} & \textbf{Aux. Info.}  & ~~\textbf{Multi-S.}~~ & ~~~\boldsymbol{$\delta_1$}$\uparrow$~~  & \textbf{REL}$\downarrow$   & \textbf{RMS}$\downarrow$ \\
        \hline
        Reg. Baseline  &  &  &  &  &  & 0.852 & 0.130 & 0.422 \\
        Reg. Baseline+  &  &  & $\surd$ &  &  & 0.870 & 0.123 & 0.403 \\
        Fix UD  & $\surd$ &  &  &  &  & 0.851 & 0.130 & 0.424 \\
        Fix SID  & $\surd$ &  &  &  &  & 0.825 & 0.145 & 0.453  \\
        Adabins~\cite{bhat2021adabins} & $\surd$ & $\surd$ &  & $\surd$ &  & 0.850 & 0.136 & 0.434\\
        \hline
        \multirow{3}{*}{\textbf{BinsFormer}}
        &$\surd$ & $\surd$ & $\surd$ &  &  & 0.878 & 0.116 & 0.397\\
        &$\surd$ & $\surd$ & $\surd$ & $\surd$ &  & 0.882 & 0.115 & 0.388\\
        &$\surd$ & $\surd$ & $\surd$ & $\surd$ & $\surd$ & \textbf{0.890} & \textbf{0.113} & \textbf{0.379} \\
        \hline
        \end{tabular} 
    }
\end{table}

% query, no aux, multi-level 1,
% query, aux (cls), multi-level 1,
% query, aux (cls), multi-level 3, 

\begin{figure}[t]
    \centering
    \footnotesize
    \begin{tabular}{l}
        \includegraphics[width=0.95\linewidth]{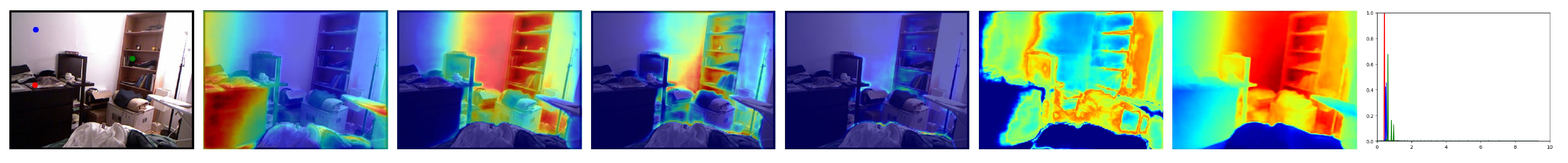} \vspace{-0.2cm}\\
        \includegraphics[width=0.95\linewidth]{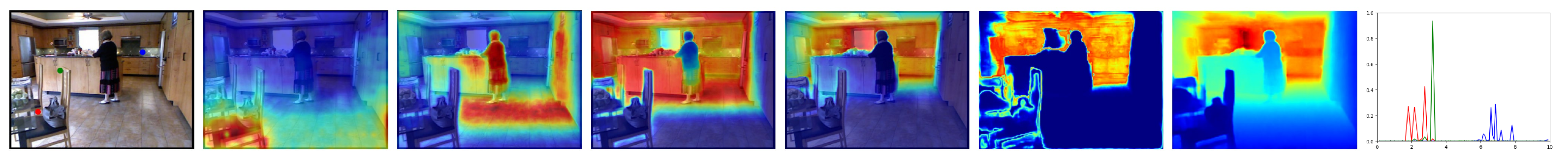} \vspace{-0.2cm}\\
        \includegraphics[width=0.95\linewidth]{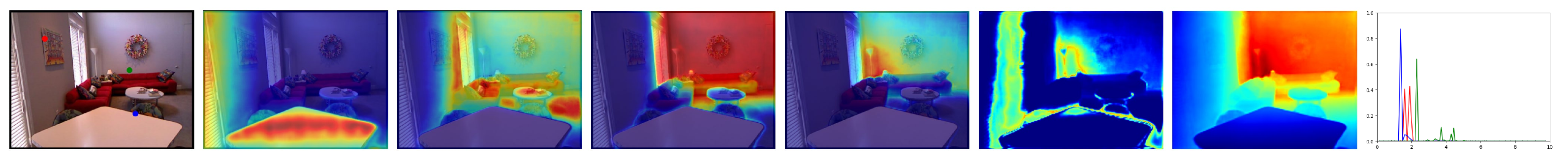} \\
        \hspace{0.03\linewidth}RGB
        \hspace{0.03\linewidth}Query 8
        \hspace{0.01\linewidth}Query 24
        \hspace{0.002\linewidth}Query 50
        \hspace{0.002\linewidth}Query 64
        \hspace{0.012\linewidth}Uncert.
        \hspace{0.03\linewidth}Depth
        \hspace{0.05\linewidth}Prob.\\
    \end{tabular}
    \caption{Visualization of different areas that queries are responsible for sensing, uncertainty maps, and probability distribution of selected points in images. We randomly select three points in the RGB images and plot their distributions in the Prob. plots, where Bins centers are presented by the small ticks upon the x-aris. Figure Query $N$ visualizes the similarity of $N^{th}$ query and the per-pixel representation $f_p$.}
    \label{fig::visual-1}
\end{figure}

\subsubsection{BinsFormer:} We first evaluate each component of BinsFormer. We start with the per-pixel regression baseline. The Reg. Baseline uses the pixel-level module of BinsFormer and directly outputs per-pixel depth predictions (\textit{i.e.,} w/o Transformer and depth estimation module). For a fair comparison, we design the Reg. Baseline+, which adds the transformer module and query embedding MLP to the Reg. Baseline but predict depth in a regression manner (\textit{i.e.,} w/o depth estimation module). Thus, Reg. Baseline+ and BinsFormer differ only in the formulation: regression \textit{v.s.} classification-regression. In terms of classification-regression methods, we compare our bins generation method with pre-defined fixed UI/SID and Adabins~\cite{bhat2021adabins}. Results presented in Tab.~\ref{tab::ablation-nyu} demonstrate the effectiveness of BinsFormer. We present bins predictions and probability distributions in Fig.~\ref{fig::visual-1}, which indicates BinsFormer can adaptively estimate suitable bins for various dynamic scenes (\textit{e.g.,} for images containing large areas of distant pixels, predicted bins rather approaches the max depth). 

Since we predict probability distribution maps for input images, it is possible to compute the measurement uncertainty~\cite{kendall2017uncertainties} for each ray by measuring the Maximum Likelihood Estimates (MLE) following~\cite{liu2019neural}. Fig.~\ref{fig::visual-1} shows a trend where uncertainty increases with distance, which has also been observed in unsupervised models that are capable of estimating uncertainty~\cite{johnston2020self}. Areas of fringes show very high uncertainty, likely attributed to the drastic variation of depth values and the lack of depth cues in these regions.

\begin{table}[t!]
    \caption{Ablation study on the NYU dataset: effect of number of bins (N) on performance. Similar to Adabins~\cite{bhat2021adabins}, we observe that performance starts to saturate as N increases above 64.}
    \centering
    % \small
    \scalebox{0.89}{%
        \begin{tabular}{|c|c|c|c|c|c|c|}
        \hline
        $\#$ \textbf{of queries}     & ~~\boldsymbol{$\delta_1$}$\uparrow$~~  & ~~\boldsymbol{$\delta_2$}$\uparrow$~~    & ~~\boldsymbol{$\delta_3$}$\uparrow$~~     & \textbf{REL}$\downarrow$   & \textbf{RMS}$\downarrow$  & \boldsymbol{$log_{10}$}$\downarrow$ \\
        \hline
        1 & \multicolumn{6}{c|}{Not Converge} \\
        \hline
        8 & 0.889 & 0.981 & 0.995 & 0.115 & 0.390 & 0.048 \\
        16 & 0.887 & 0.981 & 0.995 & 0.113 & 0.384 & 0.048 \\
        32 & 0.889 & 0.982 & 0.995 & 0.114 & 0.380 & 0.047\\
        \cellcolor{LightCyan}\textbf{64} & \cellcolor{LightCyan}\textbf{0.890} & \cellcolor{LightCyan}\textbf{0.983} & \cellcolor{LightCyan}\textbf{0.996} & \cellcolor{LightCyan}0.113 & \cellcolor{LightCyan}\textbf{0.379} & \cellcolor{LightCyan}0.047 \\
        128 & 0.889 & 0.982 & 0.994 & \textbf{0.112} & 0.380 & \textbf{0.046} \\
        256 & 0.886 & 0.980 & 0.995 & 0.115 & 0.383 & 0.048 \\
        512 & 0.883 & 0.977 & 0.992 & 0.119 & 0.393 & 0.050 \\
        \hline
        \end{tabular} 
    }
    \label{tab::ablation-nyu-nbins}
\end{table}

% densedepth baseline
% enhdepth depthformer=False, query embeddings have no use
% only test the transformer's influence

% query, aux (cls), multi-level 3

\subsubsection{Number of bins:} To study the influence of the number of bins, we train our network for various values of $N$ bins and measure the model performance. Results are presented in Tab.~\ref{tab::ablation-nyu-nbins}. The enhancing performance gained by increasing the number of queries diminishes above $N=64$. Hence, we use $N=64$ for our final model. On top of that, we exemplify different areas that queries are responsible for sensing in Fig.~\ref{fig::visual-1}. Since queries and predicted bins are one-to-one correspondences, there is a strong correlation between depth and interest area of queries.

\subsubsection{Multi-scale strategy:} 
\begin{figure}[t]
    \centering
    \footnotesize
    \begin{tabular}{l}
        \includegraphics[width=0.95\linewidth]{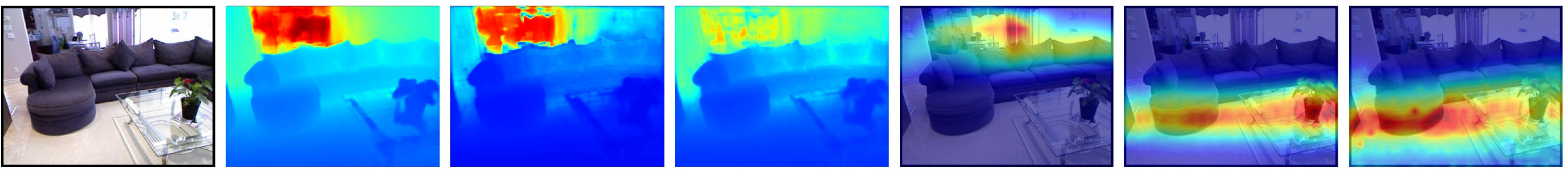} \vspace{-0.2cm}\\
        \includegraphics[width=0.95\linewidth]{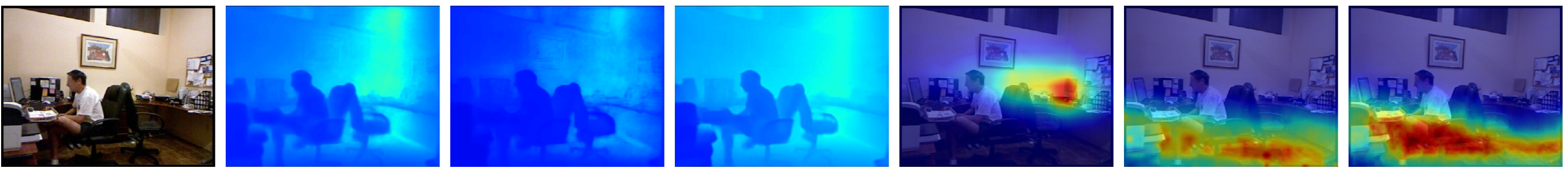} \vspace{-0.2cm}\\
        \includegraphics[width=0.95\linewidth]{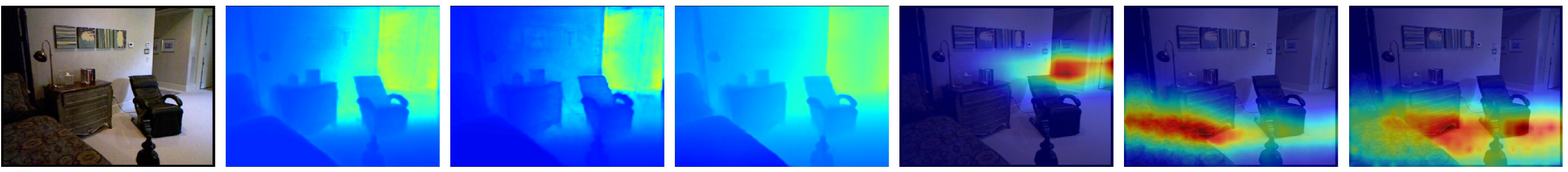} \\
        \hspace{0.035\linewidth}RGB
        \hspace{0.13\linewidth}Depth at scale 1-3
        \hspace{0.09\linewidth}Att. maps of cls. query at scale 1-3\\
    \end{tabular}
    \caption{Visualization of depth esitmation results and attention maps of the scene understanding query at the progressive multi-scale prediction refinement process.}
    \label{fig::visual-2}
\end{figure}

We then study the effectiveness of the proposed multi-scale strategy by changing the input features $\textbf{F}$ of the Transformer module. Results are presented in Tab.~\ref{tab::ablation-nyu-multilevel}. Interestingly, the performance does not always improve by adding more scale information. In detail, the performance increases significantly with the scale of the feature increasing until the number reaches three. We also provide immediate scale predictions in Fig.~\ref{fig::visual-2}, which demonstrates predictions get sharper and more accurate with the help of the proposed multi-scale refinement strategy.

\begin{table}[t!]
    \caption{Ablation study on the NYU dataset: performance of BinsFormer for different scales refinement.}
    \centering
    % \small
    \scalebox{0.89}{% 
        \begin{tabular}{|c|c|c|c|c|c|c|c|c|}
        \hline
        \boldsymbol{$f^e$}   & $\#$ \textbf{layers}  & ~~\boldsymbol{$\delta_1$}$\uparrow$~~  & ~~\boldsymbol{$\delta_2$}$\uparrow$~~    & ~~\boldsymbol{$\delta_3$}$\uparrow$~~     & \textbf{REL}$\downarrow$   & \textbf{RMS}$\downarrow$  & \boldsymbol{$log_{10}$}$\downarrow$ \\
        \hline
        % -  & 0 & - & - & - & - & - & - \\
        $f_4$  & 3 & 0.882 & 0.980 & 0.995 & 0.115 & 0.388 & 0.048 \\
        $f_3, f_4$  & 6 & 0.888 & 0.982 & 0.995 & \textbf{0.113} & 0.380 & \textbf{0.047} \\
        \cellcolor{LightCyan}\boldsymbol{$f_2, f_3, f_4$} & \cellcolor{LightCyan}\textbf{9} & \cellcolor{LightCyan}\textbf{0.890} & \cellcolor{LightCyan}\textbf{0.983} & \cellcolor{LightCyan}\textbf{0.996} & \cellcolor{LightCyan}\textbf{0.113} & \cellcolor{LightCyan}\textbf{0.379} & \cellcolor{LightCyan}\textbf{0.047} \\
        $f_1, f_2, f_3, f_4$  & 12 & 0.881 & 0.979 & 0.994 & 0.118 & 0.392 & 0.049 \\
        \hline
        \end{tabular} 
    }
    \label{tab::ablation-nyu-multilevel}
\end{table}

% when multi-level=0, no interaction between queries and features.
% It reduce to a trainable param. Specific for one dataset.

% query, aux (cls), multi-level 1-4,

\subsubsection{Auxiliary scene classification:} Results in Tab.~\ref{tab::ablation-nyu} demonstrate that the auxiliary scene understanding task can improve the model performance by introducing implicit supervision. Since the first query is responsible for classifying the environment, we visualize attention maps with the multi-scale features $F$ to investigate the effectiveness of the auxiliary scene classification. As shown in Fig.~\ref{fig::visual-2}, the classification query considers different areas of features at different scales and fully aggregates spatial information to provide hints for bins embeddings. 

\section{Conclusion}

In this paper, we propose a novel classification-regression based monocular depth estimation method, called \textbf{BinsFormer}, that incorporates a Transformer module to prediction bins in a set-to-set manner, a per-pixel module to estimate high-resolution pixel-wise representations, and a depth estimation module to aggregate information to predict final depth maps. BinsFormer can adaptively generate bins and per-pixel probability distribution for accurate depth estimation. We propose an auxiliary scene understanding task and a multi-scale prediction refinement strategy that can be seamlessly integrated into the Transformer module. These two methods further boost model performance and only introduce negligible overhead. The performance of BinFormer achieves new state-of-the-art on two popular benchmark datasets. Moreover, the generalization ability of the method was further demonstrated in cross dataset experiments.

% \clearpage
\bibliographystyle{splncs04}
\bibliography{egbib}

%%%%%%%%%%%%%%%%%% Supplementary material %%%%%%%%%%%%%%%%%%%%%%%%%%%%

\clearpage
\appendix
\section{Appendix}

\subsection{Network Structure}
\begin{figure}[t]
    \includegraphics[width=1\linewidth]{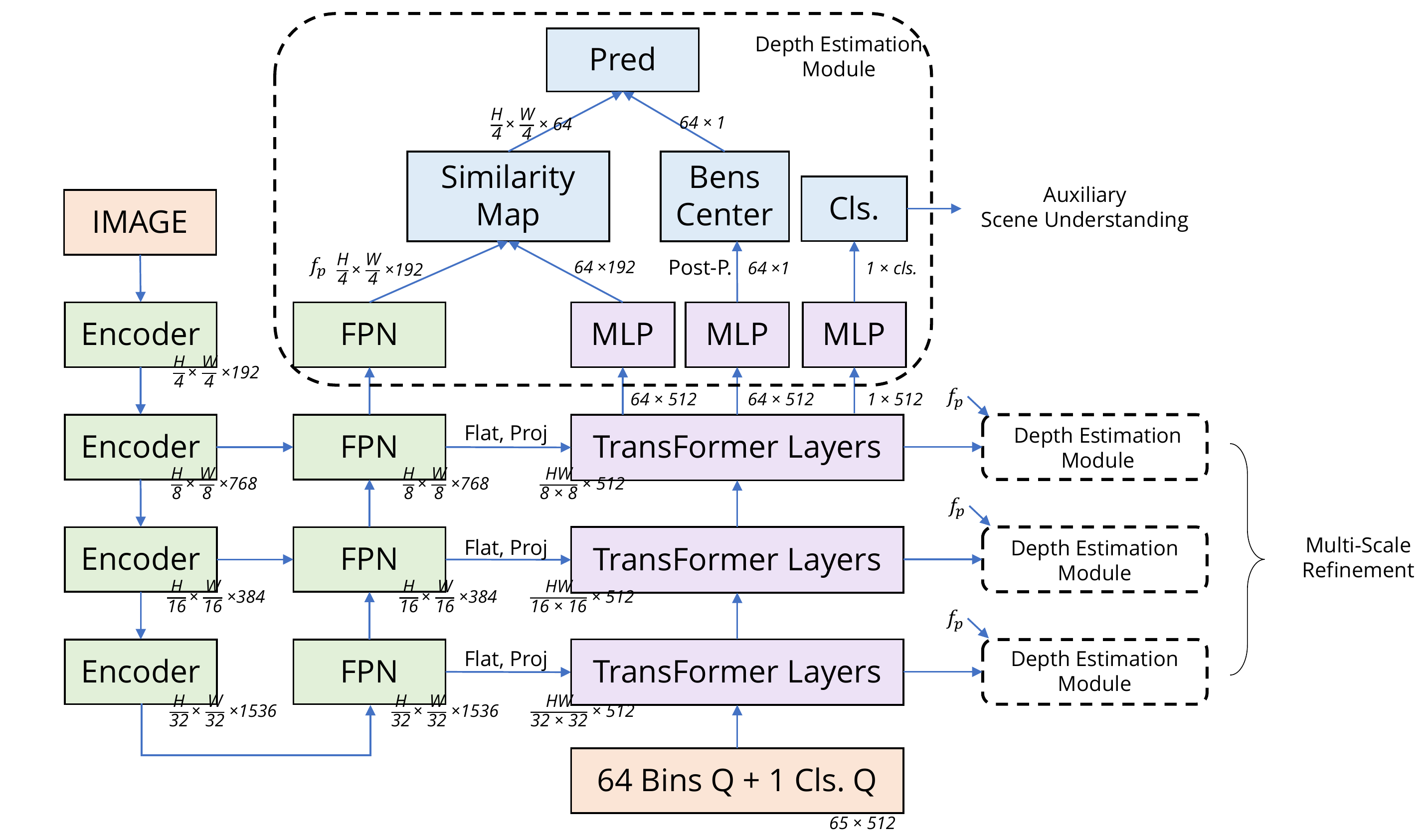}
    \caption{Network details of BinsFormer.}
    \label{fig::framework-sup}
\end{figure}

More details of our network are shown in Fig~\ref{fig::framework-sup}. The outputs of FPN are fed into the Transformer module and interact with queries. MLPs project queries into bins embeddings, bins centers, and auxiliary scene classification results. The depth estimation module aggregates the information and predicts the final depth maps. It is also applied to each scale output to achieve multi-scale refinement.

\begin{figure}[t]
    \begin{tabular}{ccc}
        \includegraphics[width=0.32\textwidth]{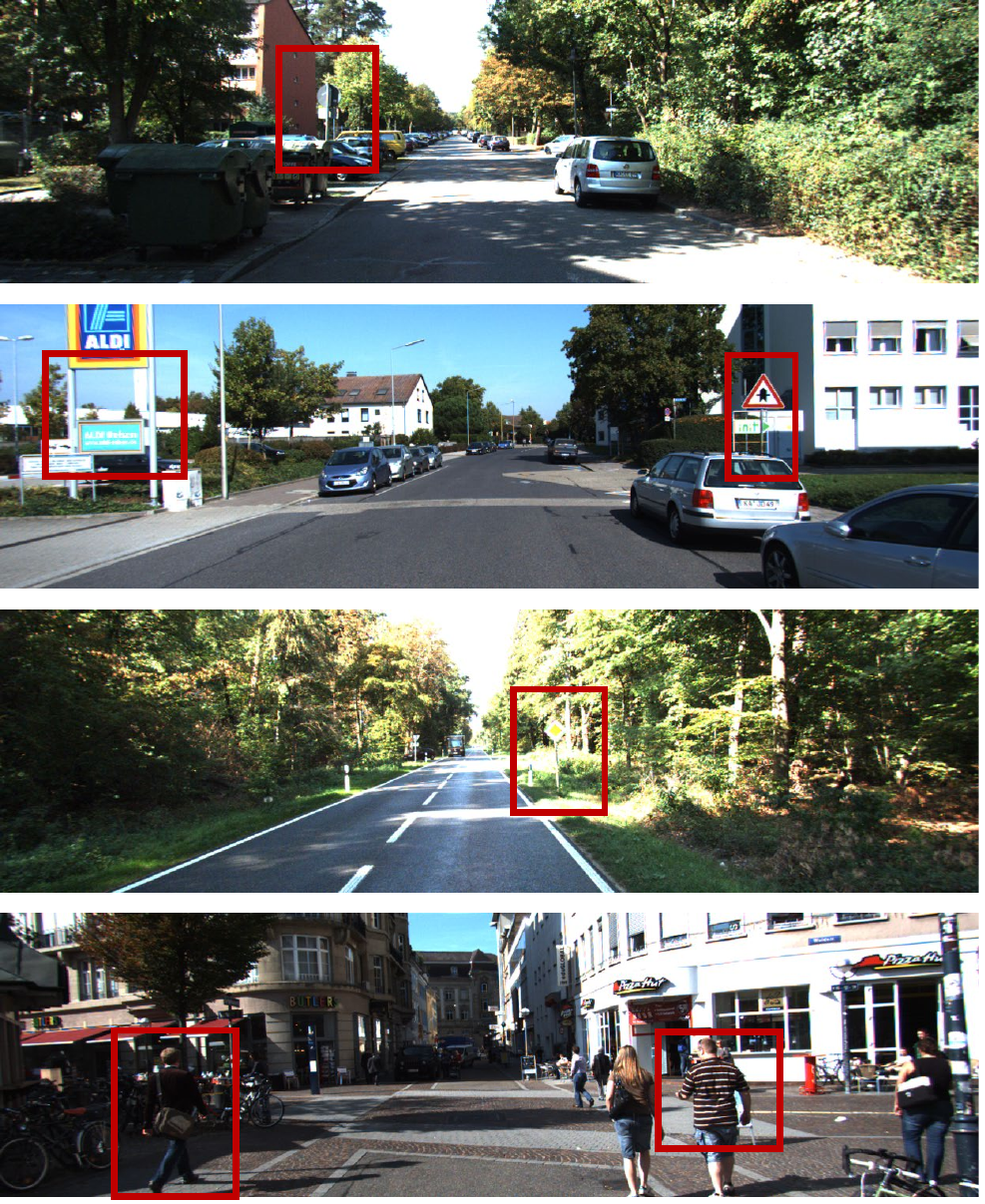}&
        \includegraphics[width=0.32\textwidth]{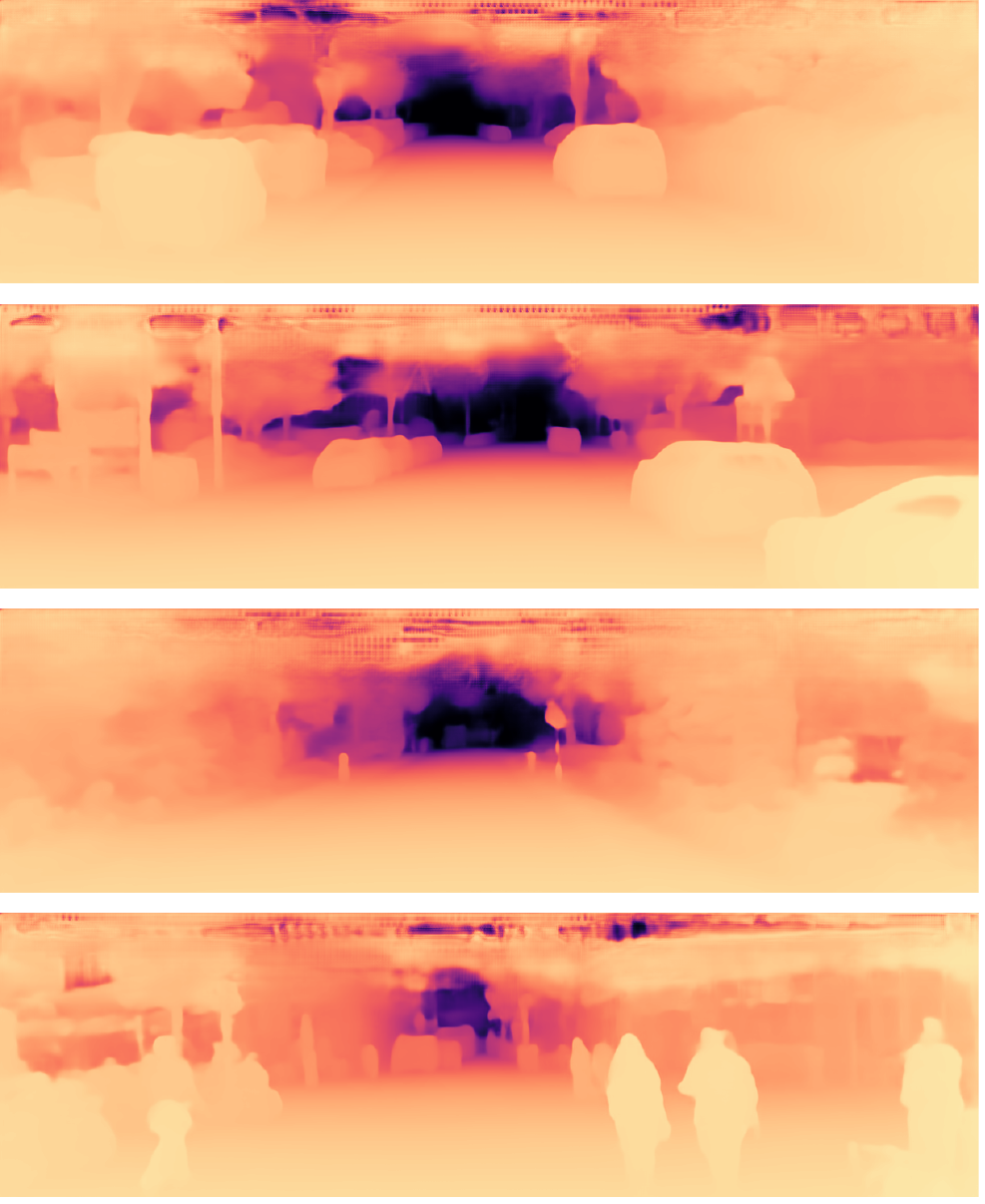}&
        \includegraphics[width=0.32\textwidth]{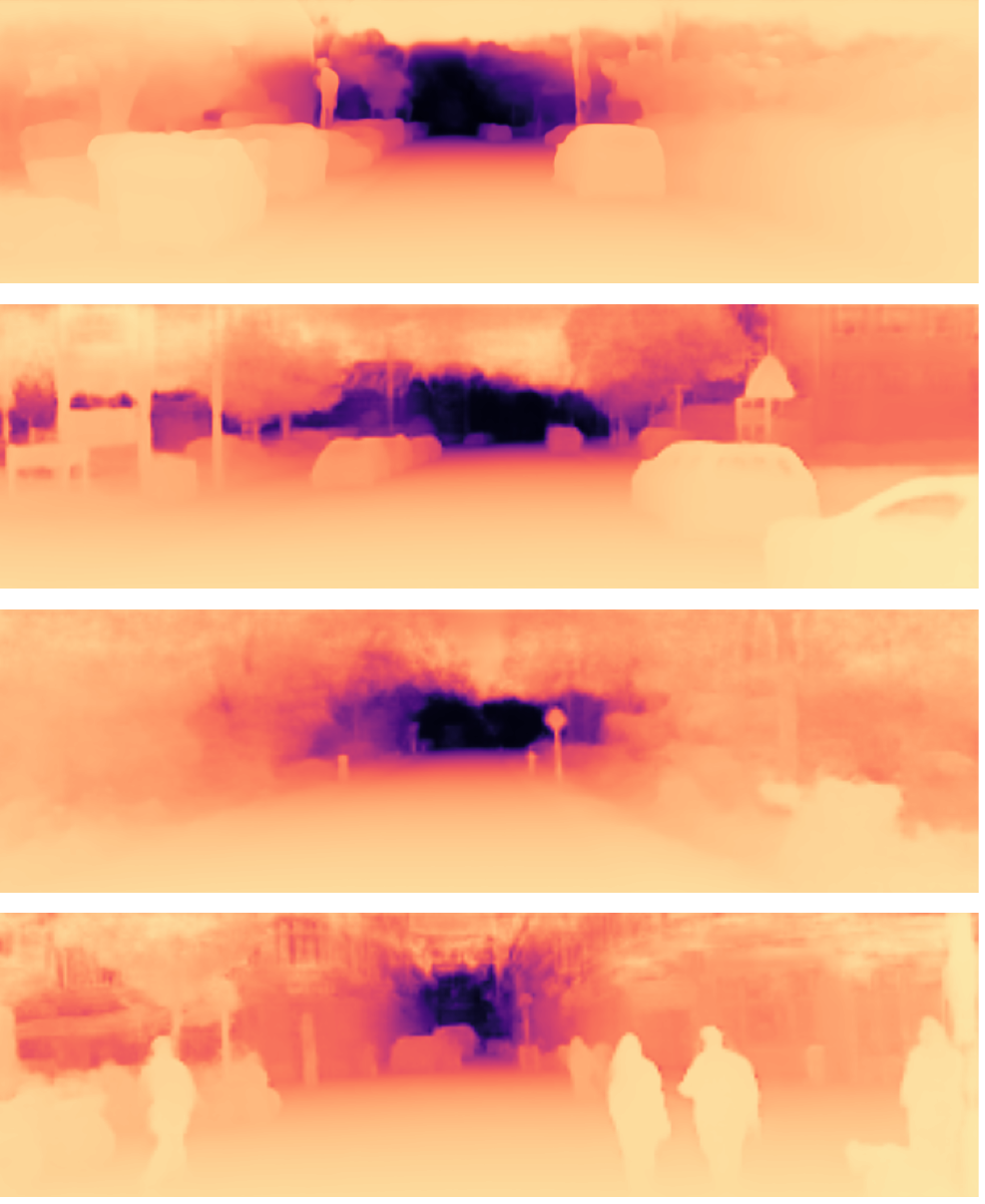}\\
        Input & BTS~\cite{lee2019bts} & BinsFormer
    \end{tabular}
    \caption{Qualitative results on KITTI dataset.}
    \label{fig::kitti}
\end{figure}

\begin{figure}[t]
    \begin{tabular}{cccc}
        \includegraphics[width=0.24\textwidth]{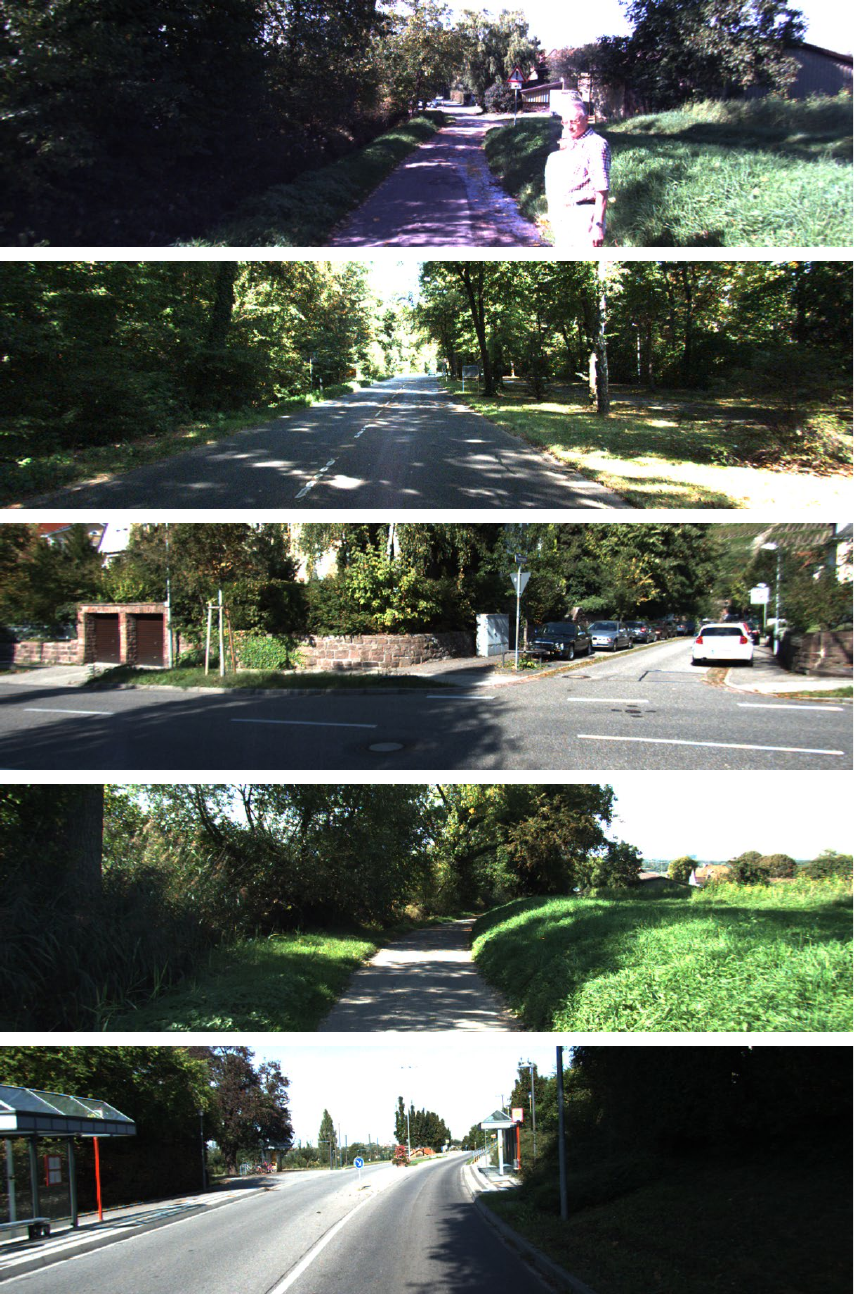}&
        \includegraphics[width=0.24\textwidth]{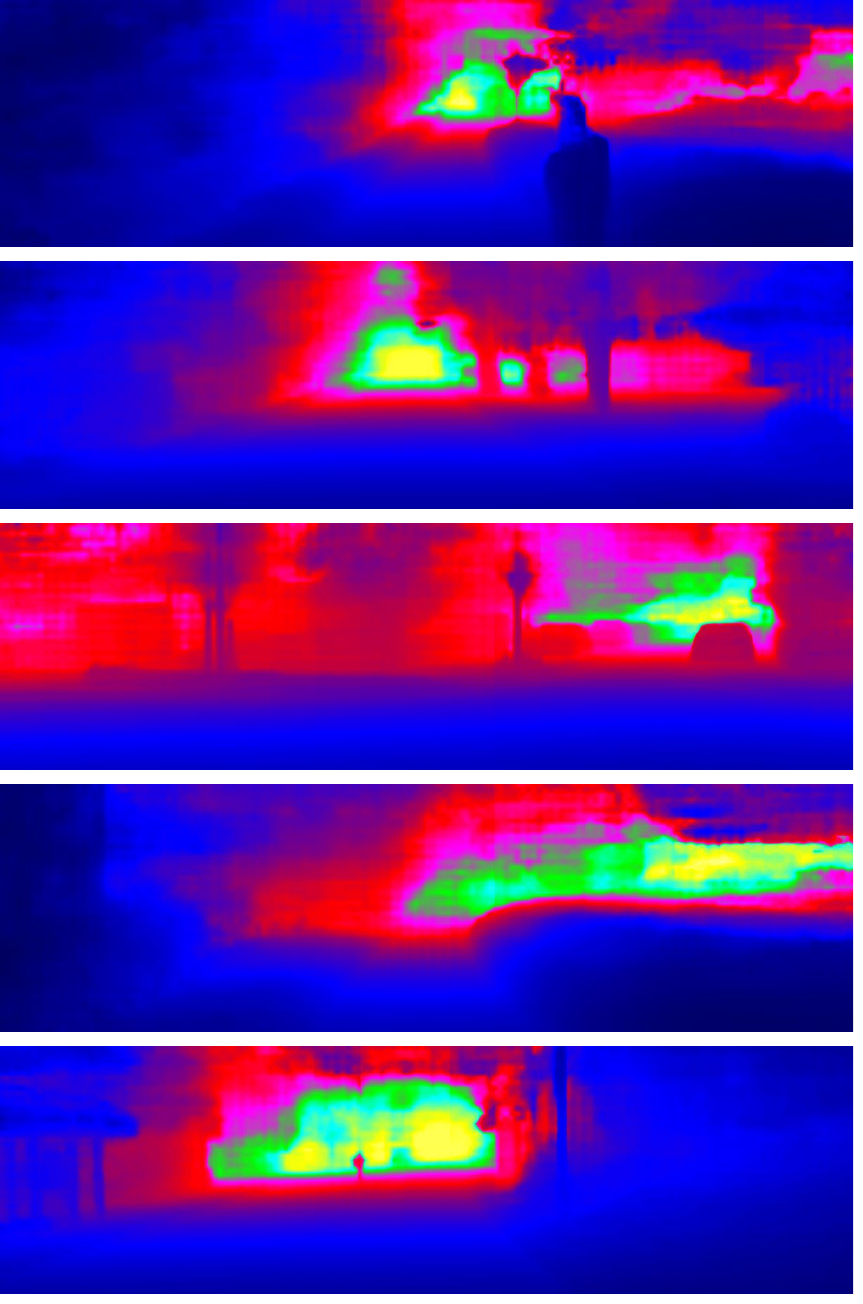}&
        \includegraphics[width=0.24\textwidth]{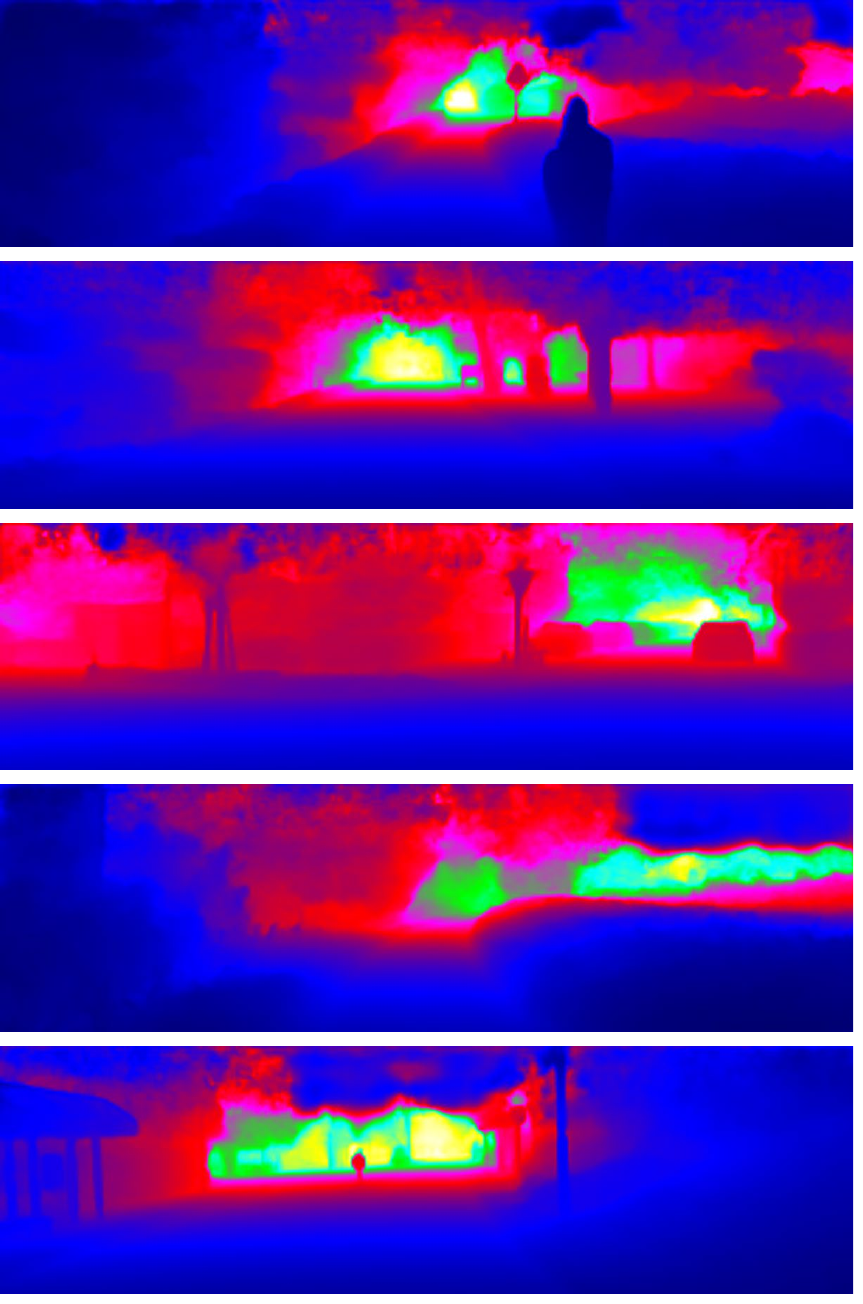}&
        \includegraphics[width=0.24\textwidth]{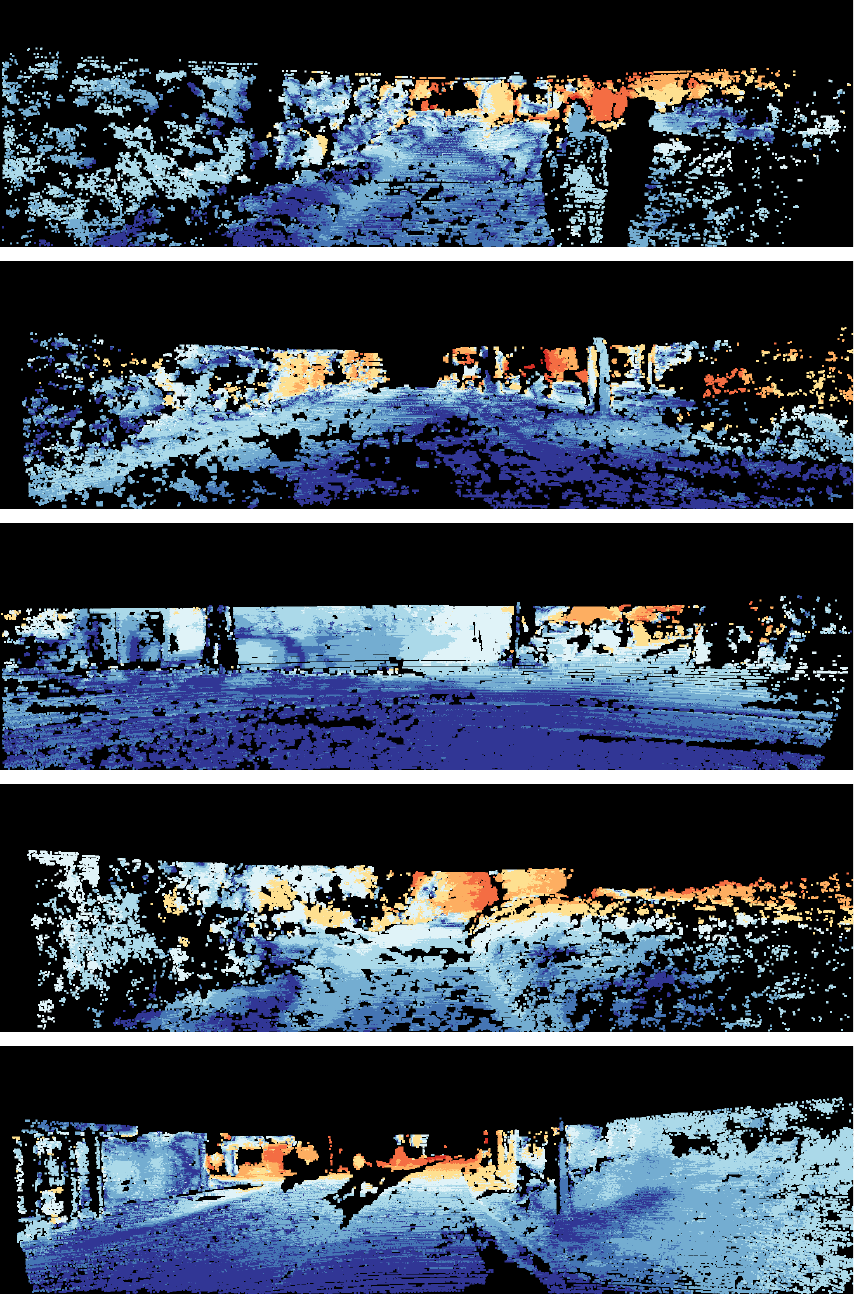}\\
        Input & DORN~\cite{fu2018dorn} & BinsFormer & Error Map
    \end{tabular}
    \caption{Qualitative results on KITTI online benchmark.}
    \label{fig::kitti-benchmark}
\end{figure}
  
\begin{figure}[t]
    \begin{tabular}{cccc}
        \includegraphics[width=0.24\textwidth]{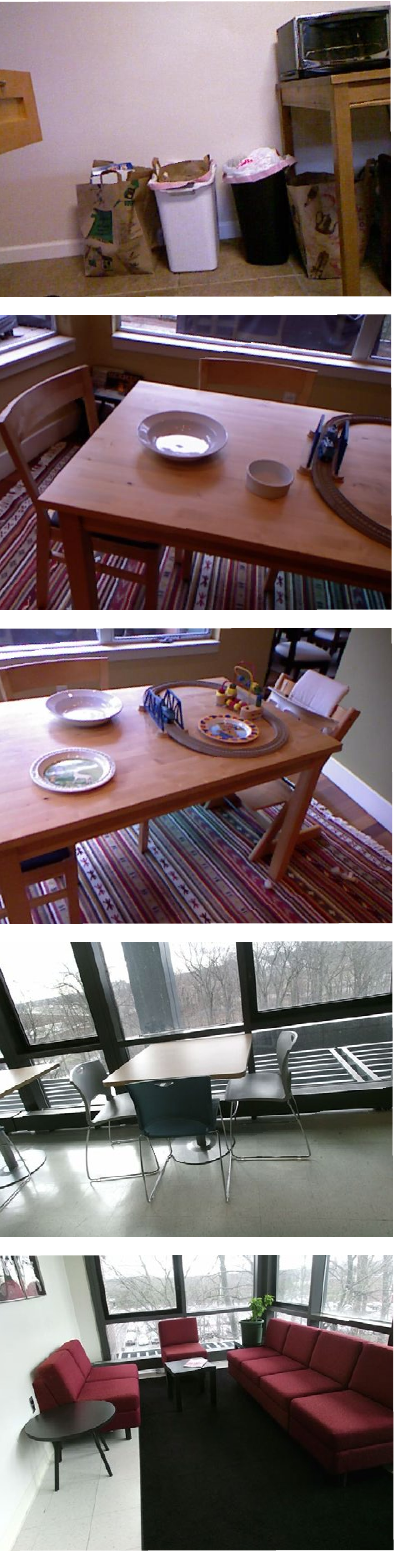}&
        \includegraphics[width=0.24\textwidth]{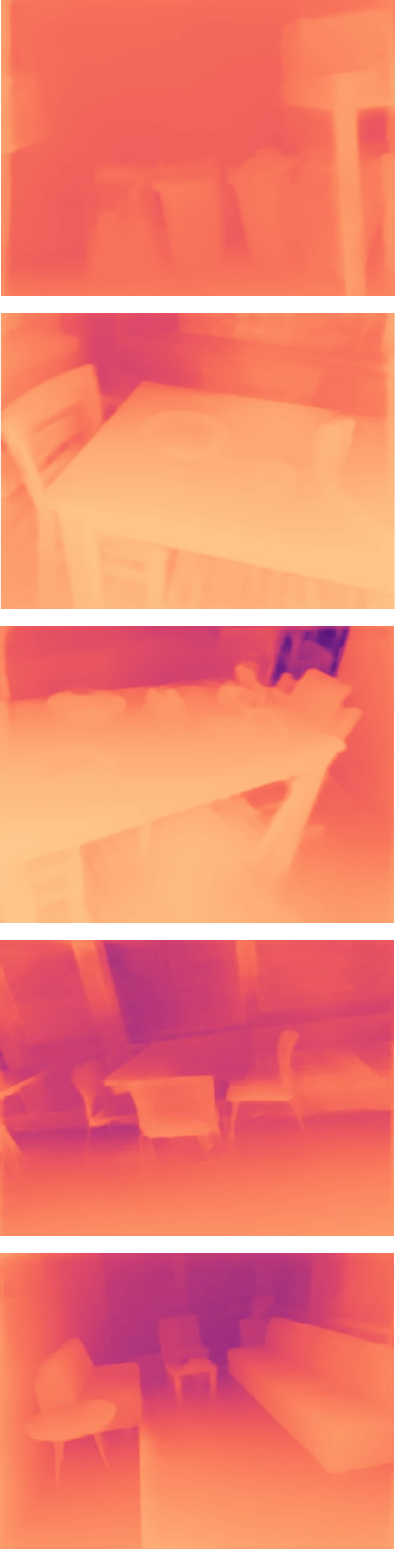}&
        \includegraphics[width=0.24\textwidth]{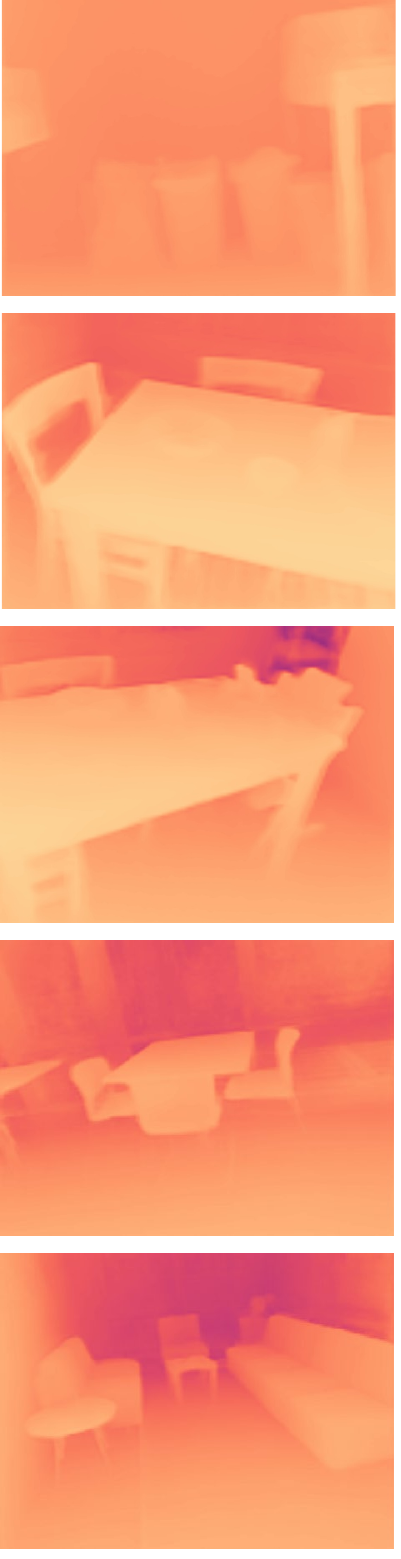}&
        \includegraphics[width=0.24\textwidth]{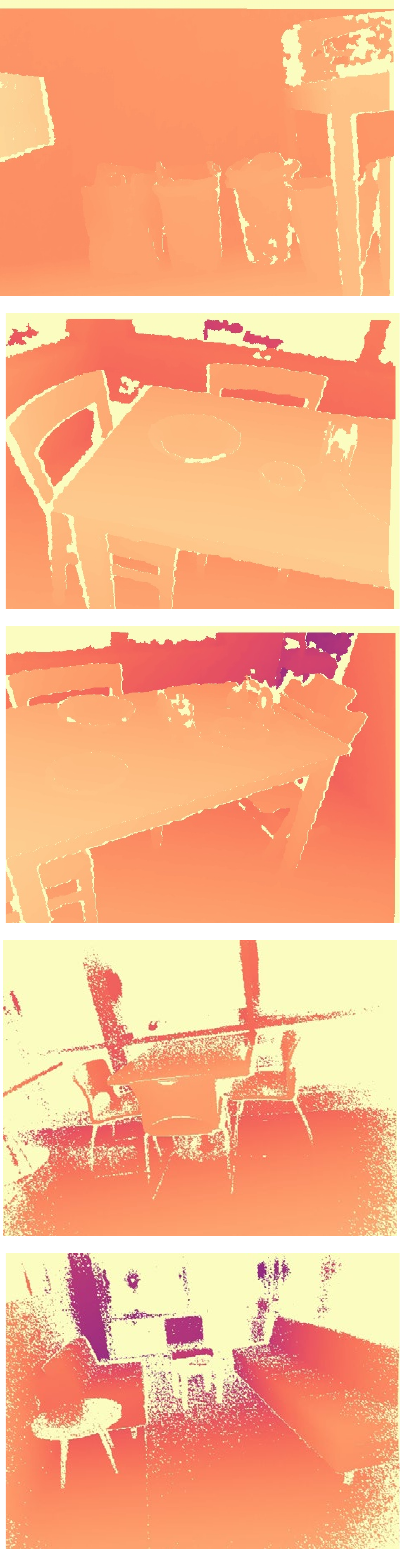}\\
        Input & Adabins~\cite{bhat2021adabins} & BinsFormer & GT
    \end{tabular}
    \caption{Qualitative results on unseen SUN RGB-D dataset.}
    \label{fig::sun-rgbd}
\end{figure}

\begin{figure}[t]
    \begin{tabular}{ccc}
        \includegraphics[width=0.32\textwidth]{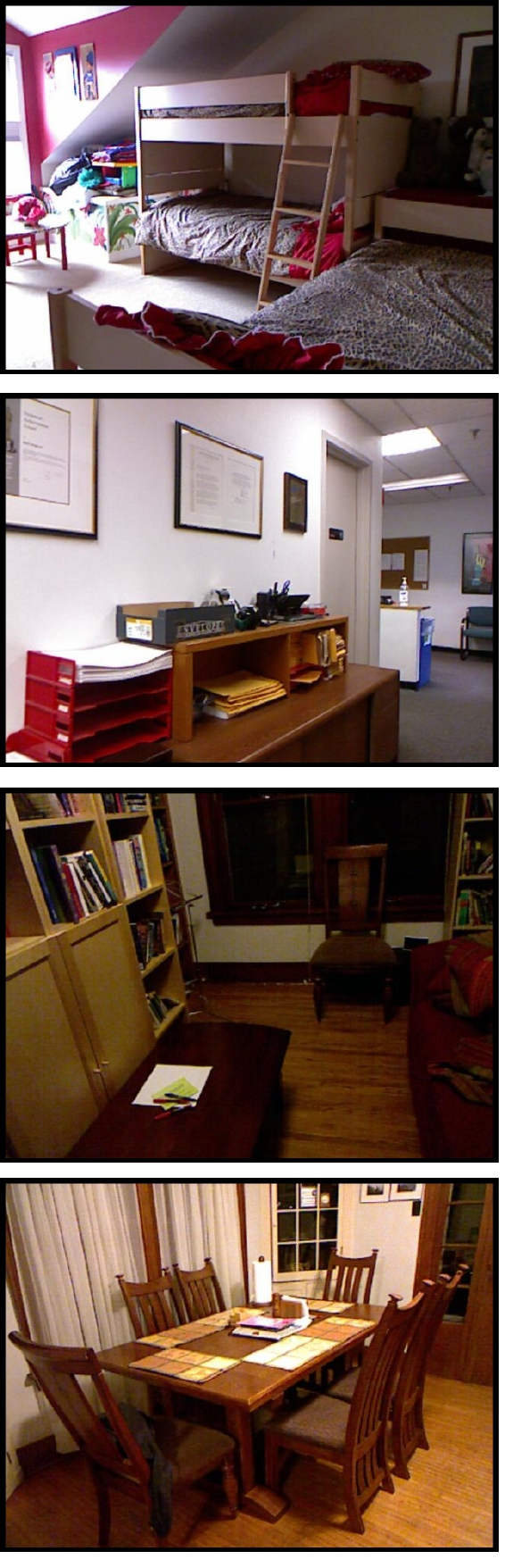}&
        \includegraphics[width=0.32\textwidth]{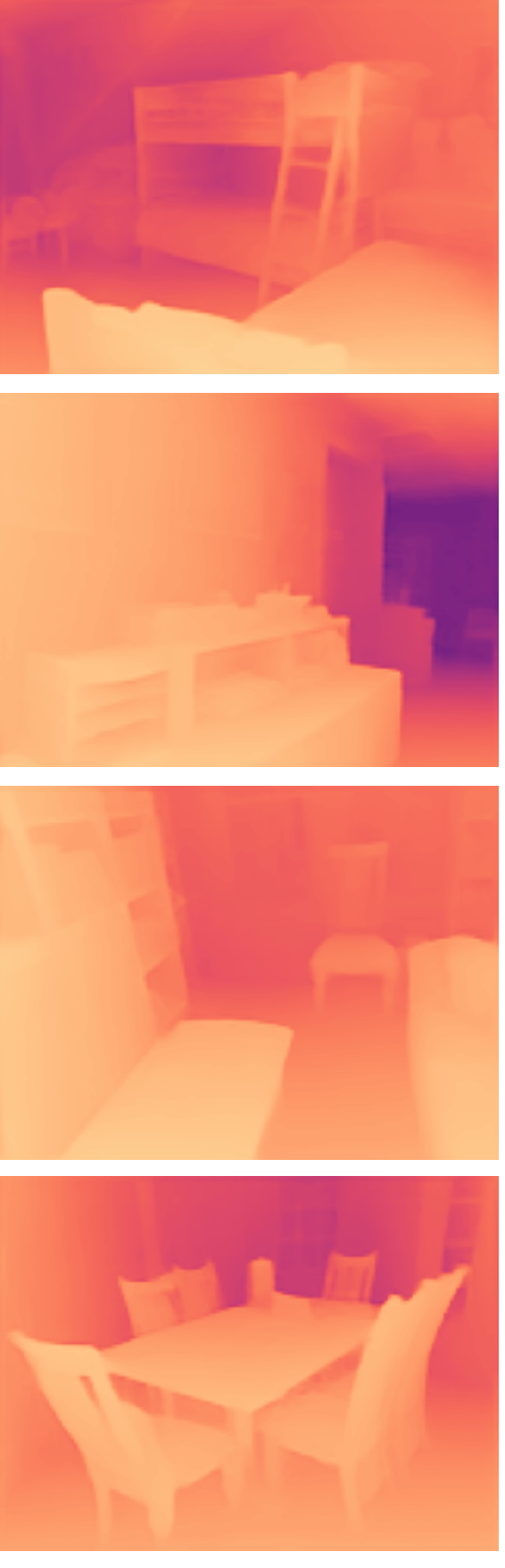}&
        \includegraphics[width=0.32\textwidth]{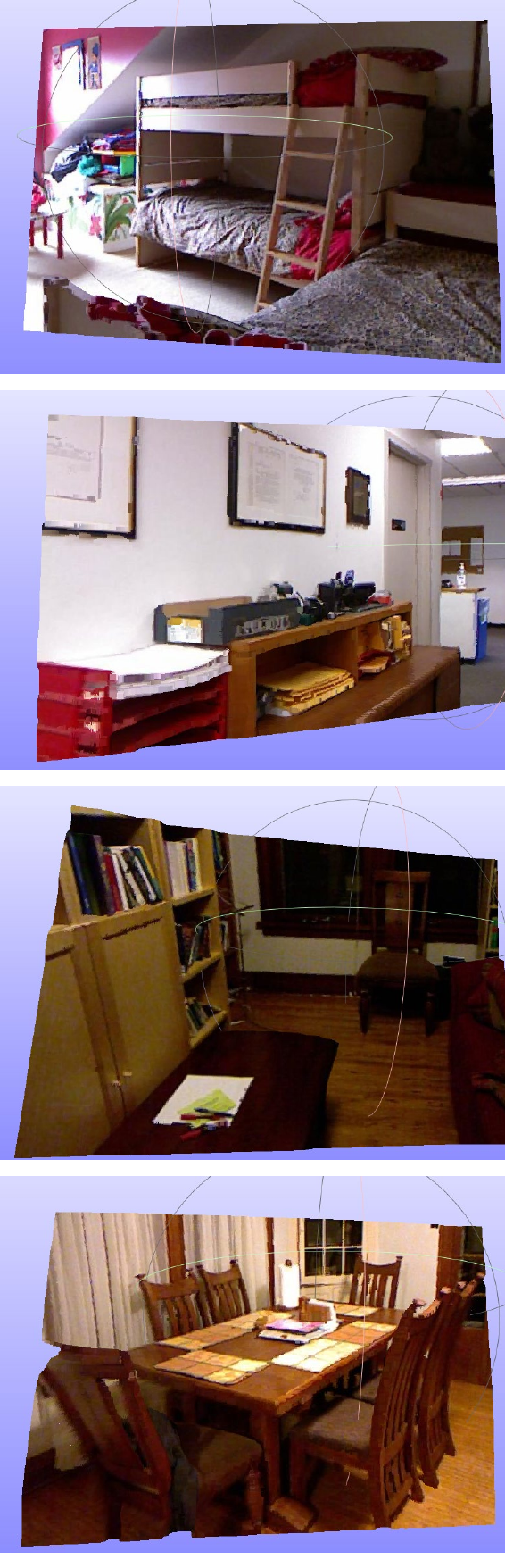}\\
        Input & Pred. Depth & Point Cloud
    \end{tabular}
    \caption{Point cloud visualization on the test set of NYUv2 dataset.}
    \label{fig::pc}
\end{figure}

\subsection{More Qualitative Results}
Here, we present more qualitative result comparisons to previous state-of-the-art methods, including BTS~\cite{lee2019bts}, Adabins~\cite{bhat2021adabins} and DORN~\cite{fu2018dorn}. Detailed results on KITTI, KITTI online benchmark, SUN RGB-D are shown in Fig.~\ref{fig::kitti}, Fig.~\ref{fig::kitti-benchmark} and Fig.~\ref{fig::sun-rgbd}, respectively. From the results, BinsFormer estimates better depth and recovers more details. Benefiting from the reasonable global bins generation, it predicts depth with an accurate overall scale.

\subsection{Point Cloud Visualization}
Monocular depth estimation aims to recover the 3D structures. To better see the 3D shape of the estimated depth map, we inv-project the image pixel back to the 3D space via the estimated depth maps. The point cloud visualizations on the test set of NYUv2 are shown in Fig.~\ref{fig::pc}, where 3D structures can be recovered satisfyingly.

\clearpage

\end{document}